\pdfoutput=1

\documentclass[11pt]{article}
\usepackage{authblk}

\usepackage[]{ACL2023}

\usepackage{times}
\usepackage{latexsym}

\usepackage[T1]{fontenc}

\usepackage[utf8]{inputenc}

\usepackage{microtype}

\usepackage{inconsolata}

\usepackage{booktabs}
\usepackage{graphicx}
\usepackage{multirow}
\usepackage{subcaption}
\usepackage{xspace}
\usepackage{kotex}
\usepackage[russian,english]{babel}

\def\secref#1{\S\ref{sec:#1}}
\def\seclabel#1{\label{sec:#1}}

\def\datasetname{\textsc{REACT}\xspace}

\title{A Federated Approach to Few-Shot Hate Speech Detection for Marginalized Communities}

\author[1,2]{\bf Haotian Ye}
\author[1]{\bf Axel Wisiorek}
\author[1,2]{\bf Antonis Maronikolakis}
\author[1,3]{\\ \bf \"Ozge Alaçam}
\author[1,2]{\bf Hinrich Sch\"utze}
\affil[1]{Center for Information and Language Processing, LMU Munich}
\affil[2]{Munich Center for Machine Learning (MCML)}
\affil[3]{Computational Linguistics, Department of Linguistics, Bielefeld University
\protect\\ \texttt{\{yehao, wisiorek, antmarakis\}@cis.lmu.de}
\protect\\ \texttt{oezge.alacam@uni-bielefeld.de}
}

\begin{document}
\maketitle

\begin{quote}
{\color{red}\textbf{Disclaimer:} This paper includes examples of hateful or offensive language used solely for illustrative purposes. These examples may be upsetting to some readers and do not represent the views or beliefs of the authors.}
\end{quote}

\begin{abstract}
Hate speech online remains an understudied issue for marginalized communities, particularly in the Global South, which includes developing societies with increasing internet penetration.
In this paper, we aim to provide marginalized communities in societies where the dominant language is low-resource with a privacy-preserving tool to protect themselves from online hate speech by filtering offensive content in their native languages.
Our contributions are twofold: 1) we release \datasetname
(\textbf{RE}sponsive hate speech datasets \textbf{A}cross \textbf{C}on\textbf{T}exts), a collection of high-quality, culture-specific hate speech detection datasets
comprising multiple target groups and low-resource languages,
curated by experienced data collectors;
2) we propose a few-shot hate speech detection approach based on federated learning (FL),
a privacy-preserving method for collaboratively training a central model
that exhibits robustness when tackling different target groups and languages.
By keeping training local to user devices, we ensure data privacy while leveraging the collective learning benefits of FL.
Furthermore, we explore personalized client models tailored to specific target groups and evaluate their performance.
Our findings indicate the overall effectiveness of FL across different target groups, and point to personalization as a promising direction.

\end{abstract}

\section{Introduction}
Combating online hate is a crucial aspect of content moderation, with prevailing solutions often relying on machine learning models trained on large-scale datasets \citep{pitenis-etal-2020-offensive, rottger-etal-2021-hatecheck, nozza-2021-exposing}.
However, these efforts and the resources required are largely limited to a few high-resource languages, such as English and German.
While multilingual hate speech datasets have been developed \citep{rottger-etal-2022-multilingual, das-etal-2022-hate-speech},
a significant portion of the world's low-resource languages and their users remain unprotected from online abuse.
A key challenge in hate speech detection lies in its inherently subjective and context-dependent nature, which varies not only at the individual level but also across cultures and regions.
The issue is exacerbated by a lack of diversity among annotators, often leading to a disconnect between those labeling the data and those directly affected by hate speech
\citep{davidson-etal-2019-racial, sap-etal-2019-risk}.
Additionally, both language and hate speech constantly evolve, with new expressions and terminology regularly emerging.

To address these challenges, we develop high-quality, culturally relevant datasets that reflect the experiences of marginalized communities.
This is achieved through a prompt-based data collection procedure, carried out by data collectors proficient in the target languages and familiar with the nuances of hate speech directed at marginalized groups within their respective contexts.
The result is \datasetname
, a set of localized, context-aware datasets containing positive, neutral, and hateful sentences across various low-resource languages.
We release \datasetname under the Creative Commons ShareAlike license (CC BY-SA 4.0).

One key limitation of current hate speech filtering solutions is their reliance on centralized, server-side processing.
In such setups, user data must be transmitted to remote servers for analysis, restricting individual control over the content being filtered.
Moreover, centralized models are less adaptable to highly specific targets, particularly in low-resource language settings.

To overcome this, we propose the use of federated learning (FL) \citep{DBLP:conf/aistats/McMahanMRHA17},
a decentralized machine learning paradigm where multiple users collaboratively train a central model without sharing raw data.
FL operates in two iterative stages: first, client devices receive the current server model and train it locally on private data;
then, updates are sent back to the server, aggregated, and used to improve the server model.
This decentralized approach not only preserves user privacy but also enables rapid adaptation to culturally specific hate speech patterns.

Our work aims to tackle the following research questions.
\textbf{RQ1}: Can zero-shot or few-shot learning effectively detect hate speech in low-resourse languages?
\textbf{RQ2}: If not, can FL bridge this performance gap?
\textbf{RQ3}: Given the specificity of hate speech, does client personalization improve over zero- or few-shot learning in low-resource settings?

\section{Related Work}
\subsection{Toxic and offensive language datasets}
Earlier efforts in the detection of toxic and offensive language, including hate speech, have contributed to the curation of diverse datasets, predominantly in English \citep{waseem-hovy-2016-hateful, DBLP:conf/www/WulczynTD17, DBLP:conf/esws/ZhangRT18}
and to a lesser extent in other high-resource languages, like German and Arabic \citep{DBLP:conf/fire/0001MMPDMP19, mulki-etal-2019-l}.
More recent work has developed datasets with more fine-grained details, such as different types of abuse \citep{sap-etal-2020-social, guest-etal-2021-expert}
and target groups \citep{grimminger-klinger-2021-hate, maronikolakis-etal-2022-listening}. 
In a related manner, \citet{DBLP:conf/aies/DixonLSTV18} and \citet{rottger-etal-2021-hatecheck} adopt a template-based data generation process to construct hate speech datasets categorized by targeted subgroups.
Recognizing the need for broader linguistic coverage,
recent initiatives have expanded data collection to include multiple languages, including low-resource ones \citep{rottger-etal-2022-multilingual, das-etal-2022-hate-speech, dementieva2024ukrainian}, which is crucial for developing robust hate speech detection systems for underrepresented languages.
Notably, \citet{DBLP:journals/corr/abs-2501-08284} introduce \textit{AfriHate}, an offensive speech dataset covering 15 low-resource languages and dialects spoken in Africa.

\subsection{Hate speech detection} 
In recent years, Transformer-based \citep{DBLP:conf/nips/VaswaniSPUJGKP17} language models have emerged as the backbone of many natural language processing tasks.
This trend extends to hate speech detection, where various Transformer-based models have been employed \citep{DBLP:conf/complexnetworks/MozafariFC19, ranasinghe-zampieri-2021-mudes, DBLP:journals/talip/RanasingheZ22},
including some pre-trained specifically to identify hate and offensive content \citep{caselli-etal-2021-hatebert, sarkar-etal-2021-fbert-neural}.

More recently, large language models (LLMs) based on Transformer architectures have demonstrated remarkable capabilities across a wide range of domains \citep{DBLP:conf/nips/BrownMRSKDNSSAA20, DBLP:conf/nips/Ouyang0JAWMZASR22, webb2023emergent}. 
Despite their effectiveness, training such models remains highly data- and resource-intensive, requiring substantial computational power and centralized datasets \citep{DBLP:journals/micro/GuptaKLTLWBW22, patel2023polca}.

\subsection{Federated learning}
Public datasets used to train language models often contain personally identifiable information (PII), raising privacy concerns as models may inadvertently memorize and expose such data \citep{DBLP:conf/nips/KimYLGYO23, DBLP:conf/sp/LukasSSTWB23}.
At the same time, the rapid development of LLMs, which require increasingly vast amounts of training data, has sparked concerns over the depletion of publicly available data.
A recent study by \citet{villalobos2022will} suggests that we may reach this data limit as early as 2026.

In this context, effectively leveraging privately held data, such as that stored on user devices, in a privacy-preserving way presents a promising potential.
Federated learning (FL) \citep{DBLP:conf/aistats/McMahanMRHA17} is a decentralized machine learning paradigm designed to preserve data privacy.
Instead of collecting user data centrally, FL enables models to be trained locally on individual devices (clients), ensuring that raw data never leaves the device.
Model update from each client are then collected and aggregated on a central server using the \texttt{FederatedAveraging} (\texttt{FedAvg}) algorithm, which computes a weighted average of received local updates.
One of the first applications of FL was in improving next-word prediction in Gboard, Google's virtual keyboard \citep{hard2018federated}.
In this setting, user interactions contributed to model improvements without exposing any actual data generated by individuals.
FL has since been applied to other privacy-sensitive domains such as finance \citep{DBLP:conf/icaif/ByrdP20} and medicine \citep{sheller2020federated}.
Despite its potential, FL has only recently begun to be explored in the context of hate speech detection.
\citet{gala-etal-2023-federated} and \citet{zampieri2024federated} apply FL on public offensive speech datasets and benchmarks, demonstrating its feasibility for content moderation.
Additionally, \citet{singh-thakur-2024-generalizable} explore FL to detect hate speech in various Indic languages, showing its relevance for low-resource contexts.
In contrast to these approaches, we investigate the use of FL for few-shot hate speech detection in low-resource settings, where annotated data is extremely limited.
We further explore personalized FL to enhance adaptability to specific target groups.

\subsection{Personalized FL}
The standard FL framework assumes that client data is independently and identically distributed (i.i.d.).
In scenarios where client data is highly heterogeneous (non-i.i.d.), traditional FL may suffer from degraded performance and slow convergence due to \textit{client drift} \citep{DBLP:conf/icml/KarimireddyKMRS20, DBLP:conf/iclr/LiHYWZ20}.
In the context of hate speech detection, clients may represent marginalized or underrepresented groups whose data characteristics differ significantly from the majority.
Personalized FL offers a potential solution by allowing model customization at the client level, better addressing group-specific sociolinguistic patterns.
Additionally, it further enhances privacy by limiting the amount and type of information shared with the central server.
A straightforward approach to client personalization is FedPer \citep{arivazhagan2019federated}, which decouples the client model into base (shared) and personalized layers.
This architecture enables clients to retain parameters tailored to their local data while still contributing to the server model.
Following this approach, we apply personalized FL to integrate local adaptations with selective information sharing.

\section{\datasetname Dataset} \seclabel{react_dataset}
\begin{table*}
    \centering
    \resizebox{\textwidth}{!}{%
    \begin{tabular}{ll|rrrr|rrrr|rrrr|r}
 \toprule
        language & target & \multicolumn{4}{c}{positive} & \multicolumn{4}{c}{neutral} & \multicolumn{4}{c}{hateful} & total \\
        &  & \multicolumn{2}{c}{P+} & \multicolumn{2}{c}{P-} & \multicolumn{2}{c}{P+} & \multicolumn{2}{c}{P-} & \multicolumn{2}{c}{P+} & \multicolumn{2}{c}{P-} &  \\

        \midrule
        \multirow{2}{*}{Afrikaans} & Black people & 338 & (16.6\%) & 338 & (16.6\%) & 338 & (16.6\%)  & 338 & (16.6\%) & 338 & (16.6\%) & 338 & (16.6\%) & \textbf{2028} \\
                                   & LGBTQ & 197 & (19.3\%) & 174 & (17.1\%) & 169 & (16.6\%) & 150 & (14.8\%) & 174 & (17.1\%) & 152 & (14.9\%) & \textbf{1016} \\
        \cmidrule{3-15}
        \multirow{2}{*}{Ukrainian} & Russians & 300 & (16.6\%) & 300 & (16.6\%) & 300 & (16.6\%) & 300 & (16.6\%) & 300 & (16.6\%) & 300 & (16.6\%) & \textbf{1800} \\
                                   & Russophones & 200 & (16.6\%) & 200 & (16.6\%) & 200 & (16.6\%) & 200 & (16.6\%) & 200 & (16.6\%) & 200 & (16.6\%) & \textbf{1200} \\
        \cmidrule{3-15}
        \multirow{2}{*}{Russian} & LGBTQ & 90 & (11.7\%) & 164 & (21.2\%) & 102 & (13.2\%) & 136 & (17.6\%) & 137 & (17.7\%) & 143 & (18.5\%) & \textbf{772} \\
                                 & War victims & 158 & (8.1\%) & 157 & (8.1\%) & 194 & (9.9\%) & 260 & (13.3\%) & 542 & (27.7\%) & 649 & (33.1\%) & \textbf{1960} \\
        \cmidrule{3-15}
        Korean & Women & 214 & (16.5\%) & 210 & (16.2\%) & 206 & (15.9\%) & 221 & (17.1\%) & 245 & (18.9\%) & 198 & (15.3\%) & \textbf{1294} \\
        \bottomrule
    \end{tabular}
    }
    \caption{Number of collected sentences with their percentage across six categories of each dataset. P+: with profanity, P-: without profanity. In total, the data covers seven distinct target groups in eight languages.
    }
    \label{tab:data_table_big}
\end{table*}

We release a localized hate speech detection dataset for several marginalized groups in regions where low-resource languages are predominantly used.
We name this dataset \datasetname
(\textbf{RE}sponsive hate speech datasets \textbf{A}cross \textbf{C}on\textbf{T}exts).
To construct the dataset, we recruit data collectors who are either native or highly proficient in the target language and have deep familiarity with the sociocultural nuances and contexts of hate speech in the respective countries.
\datasetname comprises data on six target groups--Black people, LGBTQ, Russians, Russophone Ukrainians, Ukrainian war victims, and women--across four languages: Afrikaans, Korean, Russian, and Ukrainian.

Each dataset is organized into six categories based on the sentiment polarity (positive, neutral, hateful) and the presence or absence of profanity, which includes vulgar or obscene language such as swear words.
We collect data both with and without profanity within each polarity category to minimize the association of profanity with hateful content.

For each of the six categories, data collectors receive a prompt formatted as follows:

\begin{quote}
    Provide [\texttt{polarity}] text in [\texttt{target language}] about the [\texttt{target group}] [using/without using] profanity.
\end{quote}

Further details on the data collection procedure are provided in \secref{annotation_details}.
Table \ref{tab:data_table_big} shows the number of sentences collected for each category across all datasets.
Most datasets are balanced across categories and contain around 1000-2000 sentences related to the target groups.

\paragraph{Data source.} Data is collected predominantly from social media platforms like Facebook\footnote{\url{https://www.facebook.com}} and X (formerly Twitter)\footnote{\url{https://x.com}}, as well as local online forums, news articles, and comment sections.
Additional sources include books and text corpora, such as Common Crawl\footnote{\url{https://commoncrawl.org}}.
In some cases, data collectors generate synthetic examples inspired by observed hate speech patterns, either from scratch or based on similar content from other sources.
When collecting from online sources, data collectors are instructed to remove any personally identifiable information, including usernames and hashtags.
Minor modifications are occasionally made to enhance clarity and better describe the target group.
In addition, a portion of the data (under 20\% for most datasets) is generated using AI tools such as ChatGPT\footnote{\url{https://chatgpt.com}} and subsequently reviewed and refined by data collectors to ensure realism and consistency with the category (details in \secref{generated_data}).

\paragraph{Cross-annotation.}
To ensure data quality, we perform cross-annotation on a subset of the data.
Specifically, we sample sentences from each of the six categories and have them annotated by an additional native speaker of the language (details in \secref{annotation_details}).

\section{Hate speech detection experiments}
To implement federated learning (FL) using our collected data, we use the Flower framework\footnote{\url{https://flower.ai}}, chosen for its simplicity and flexibility.
FL at scale typically involves a central server connected with multiple client nodes, each operating on a user's device.
Flower supports the simulation of this setup by enabling the creation of virtual clients on a single machine, allowing us to conduct controlled FL experiments without relying on real user devices.

For our hate speech detection experiments in low-resource settings,
we focus on four language-target group combinations. These include: Afrikaans - Black people (\texttt{afr-black}), Afrikaans - LGBTQ (\texttt{afr-lgbtq}), Russian - LGBTQ (\texttt{rus-lgbtq}), and Russian - war victims (\texttt{rus-war}).

\subsection{Models}
Federated learning is commonly constrained by the large communication overhead between clients and the server, where even a small amount of transmitted data may burden the bandwidth \citep{MLSYS2019_7b770da6}.
In addition, smaller models offer greater flexibility, as they can be deployed on devices with varying computational capacities \citep{hard2018federated}.
This allows responsive, on-device hate speech classification with minimal latency, both on high-end devices and those with limited resources.

Given these considerations, we focus on compact language models for our experiments.
We evaluate a total of seven models, including four multilingual models: multilingual BERT (mBERT) \citep{devlin-etal-2019-bert}, multilingual DistilBERT (Distil-mBERT) \citep{Sanh2019DistilBERTAD}, multilingual MiniLM \citep{DBLP:conf/nips/WangW0B0020}, and XLM-RoBERTa (XLM-R) \citep{conneau-etal-2020-unsupervised}.
We also include three models without explicit multilingual pre-training: DistilBERT, ALBERT \citep{DBLP:conf/iclr/LanCGGSS20}, and TinyBERT \citep{jiao-etal-2020-tinybert}.

Comprehensive results for all seven models are provided in \secref{model_selection}.
Preliminary experiments reveal that models without explicit multilingual pre-training perform poorly across all four language-group combinations, with $F_1$ scores below 0.50 in most cases.
Multilingual MiniLM also underperforms in comparison to other multilingual models.
In contrast, mBERT and Distil-mBERT consistently achieve the highest performance ($F_1$ scores of 0.70 and 0.72 respectively on the best-performing client models).
Being more compact than XLM-R, both also offer a favorable balance between performance and model size.
Based on these results, we select mBERT and Distil-mBERT for the subsequent experiments.

\subsection{Federated learning}

    

Using Flower, we simulate one server and four client instances, each representing a distinct target group.
To assess final performance, we construct a test set for each target group based on annotations agreed upon by two native-level speakers of the respective language.
Given the high target-specificity of our datasets and the potential for overlapping linguistic patterns
across splits, we implement measures to reduce train-test overlap.
Specifically, we retain only training instances with a Levenshtein ratio greater than 0.5 with test data.
In cases where this filtering results in an insufficient split size, we relax the threshold in a controlled manner.
Further details are provided in \secref{dev_train_selection}.
To address \textbf{RQ1} and \textbf{RQ2},
we evaluate client models in both zero-shot and few-shot settings, fine-tuning them with 3, 9, and 15 sentences per target group to simulate extremely low-resource settings.
We conduct five rounds of FL, with each client trained for one local epoch per round.
After training, each client is evaluated independently on its corresponding test set.
Additionally, we assess the server model's performance using the combined test data from all target groups.
All results are reported using the macro-$F_1$ score, averaged over five different random seeds.

\subsection{Client personalization}
A core objective of this work is to support personalized hate speech detection tailored to the specific needs of individual target groups.
In line with this and to investigate \textbf{RQ3}, we implement two personalization methods during the FL process.

\paragraph{FedPer.}
FedPer, introduced by \citet{arivazhagan2019federated}, personalizes client models by making the final layers private, sharing only updates to the base (non-private) layers.
$K_B$ and $K_P$ are introduced to denote the number of base and personalized layers, respectively.
Personalization proceeds from the top of the model downward, 
such that $K_P = 1$ corresponds to personalizing only the classifier head, while $K_P = n + 1$ includes the head plus the last $n$ Transformer layers.

Following \citet{arivazhagan2019federated}, we test $K_P \in \{1, 2, 3, 4\}$ for mBERT and Distil-mBERT.
We exclude the server model from evaluation as key parameters, most importantly those of the classifier head, are client-specific and not updated centrally, making server-side performance uninformative.

\paragraph{Adapters.}
A growing body of research has explored incorporating annotators' demographics and preferences \citep{kanclerz-etal-2022-ground, fleisig-etal-2023-majority, hoeken_etal_2024_hatewic},
or even gaze features of the users \citep{alacam-etal-2024-eyes} into annotations to better capture subjectivity.
Inspired by this line of work, we introduce a small number of trainable parameters in the form of adapters \citep{DBLP:conf/icml/HoulsbyGJMLGAG19} between each pair of Transformer blocks, which serve as client-specific parameters.
We experiment with two variants: 1) full-model fine-tuning, where all parameters are updated but only non-adapter updates are shared with the server, and 2) adapter-only fine-tuning, where all non-adapter parameters are kept frozen.
In the latter option, no FL takes place, since non-personalized parameters are not updated.
As with FedPer, we exclude the server model from evaluation.

\subsection{Baseline}
To evaluate the effectiveness of FL across different target groups, we establish a standard few-shot fine-tuning baseline, where each model is trained individually on a single target group using the same data and parameters.
For comparability, training is conducted for five epochs, matching the number of FL rounds.
In addition, we evaluate performance using the Perspective API\footnote{\url{https://perspectiveapi.com}}, a widely used tool for only toxic speech filtering.
Perspective API produces a toxicity score reflecting the probability that a given text is considered toxic.
However, the classification outcome is highly sensitive to the selected toxicity threshold, and prior studies have shown that the API can exhibit biases, particularly with unfamiliar or culturally specific language use \citep{DBLP:conf/chi/HuaNR20, DBLP:journals/csur/GargMS023, DBLP:journals/corr/abs-2312-12651}.
For this reason, we report results using two toxicity thresholds of 0.7 and 0.9 according to the API's recommended range.

\section{Results}

\begin{figure*}[htbp]
    \centering
    \begin{subfigure}[b]{0.32\textwidth}
        \centering
        \includegraphics[width=\textwidth]{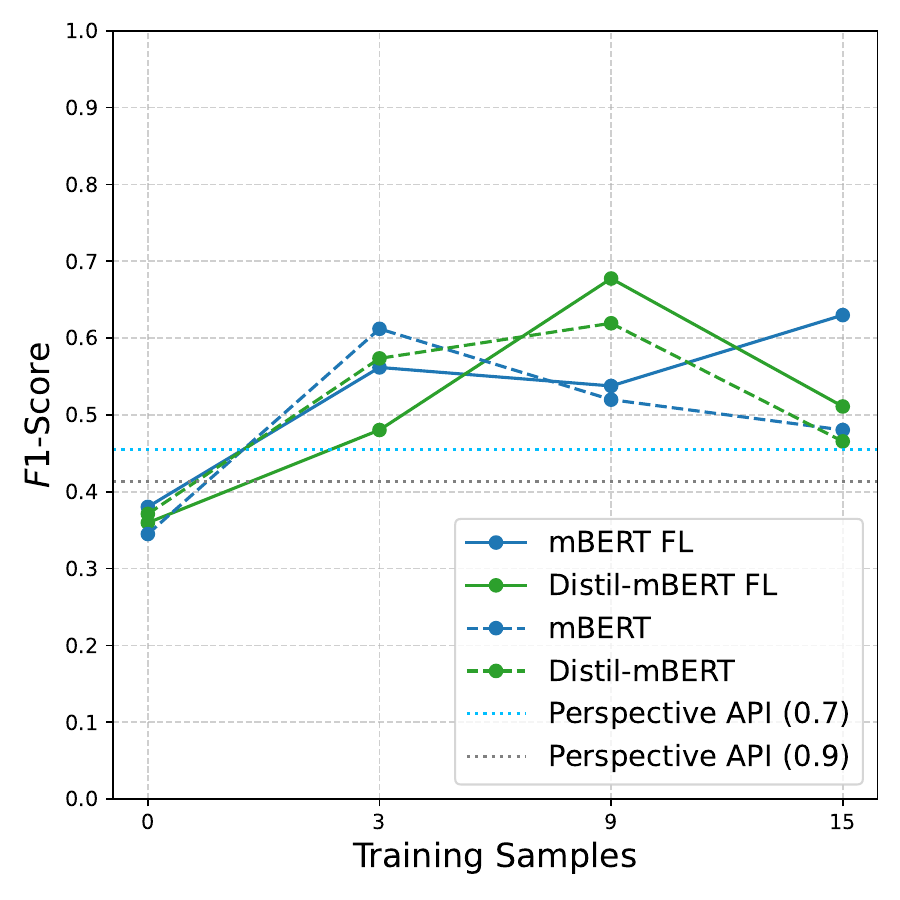}
        \caption{\texttt{afr-black}}
    \end{subfigure}
    \begin{subfigure}[b]{0.32\textwidth}
        \centering
        \includegraphics[width=\textwidth]{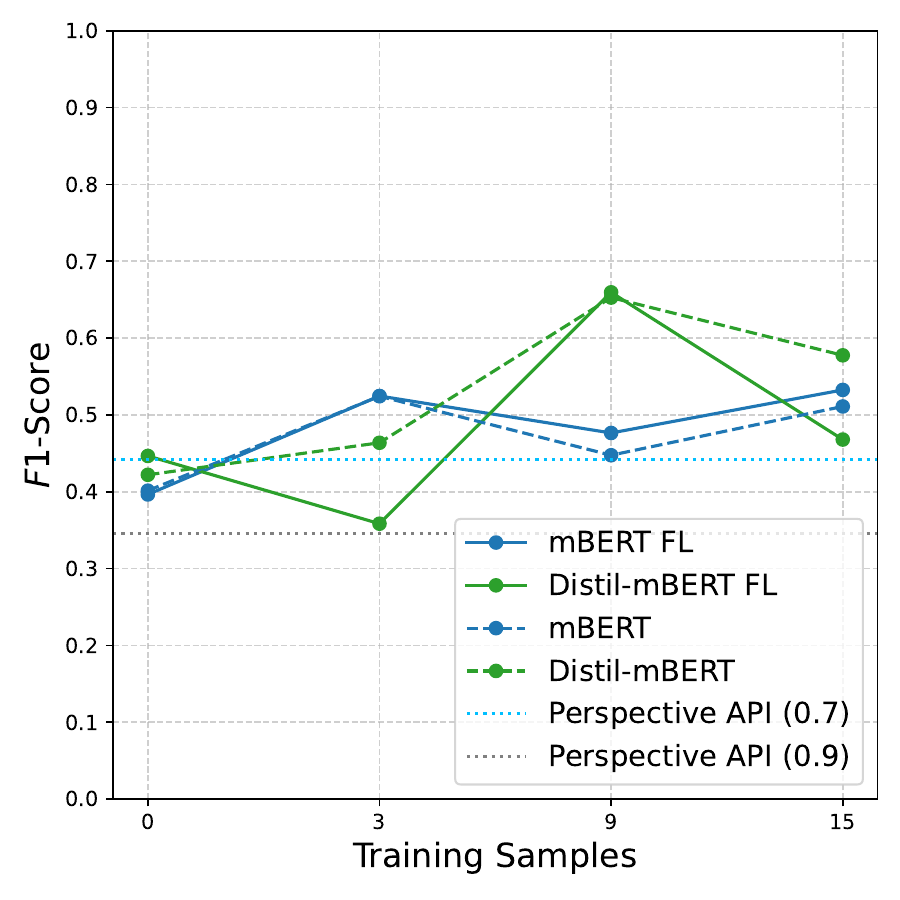}
        \caption{\texttt{afr-lgbtq}}
    \end{subfigure}
    
    \vspace{0.5cm}
    \begin{subfigure}[b]{0.32\textwidth}
        \centering
        \includegraphics[width=\textwidth]{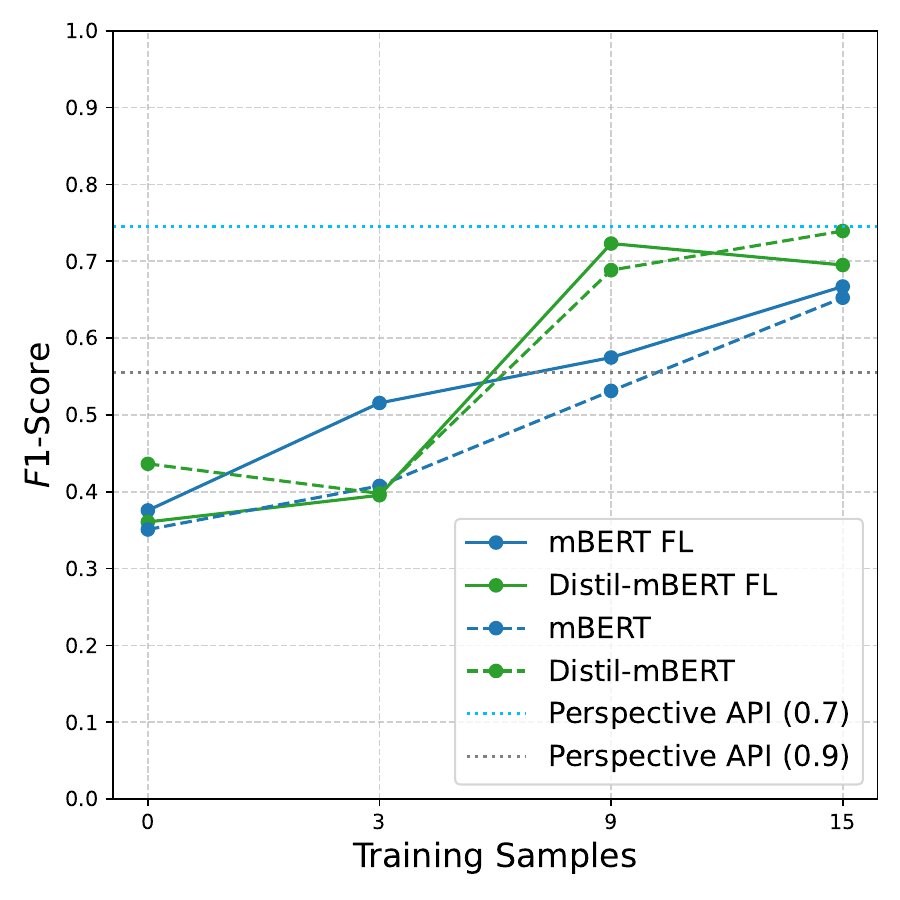}
        \caption{\texttt{rus-lgbtq}}
    \end{subfigure}
    \begin{subfigure}[b]{0.32\textwidth}
        \centering
        \includegraphics[width=\textwidth]{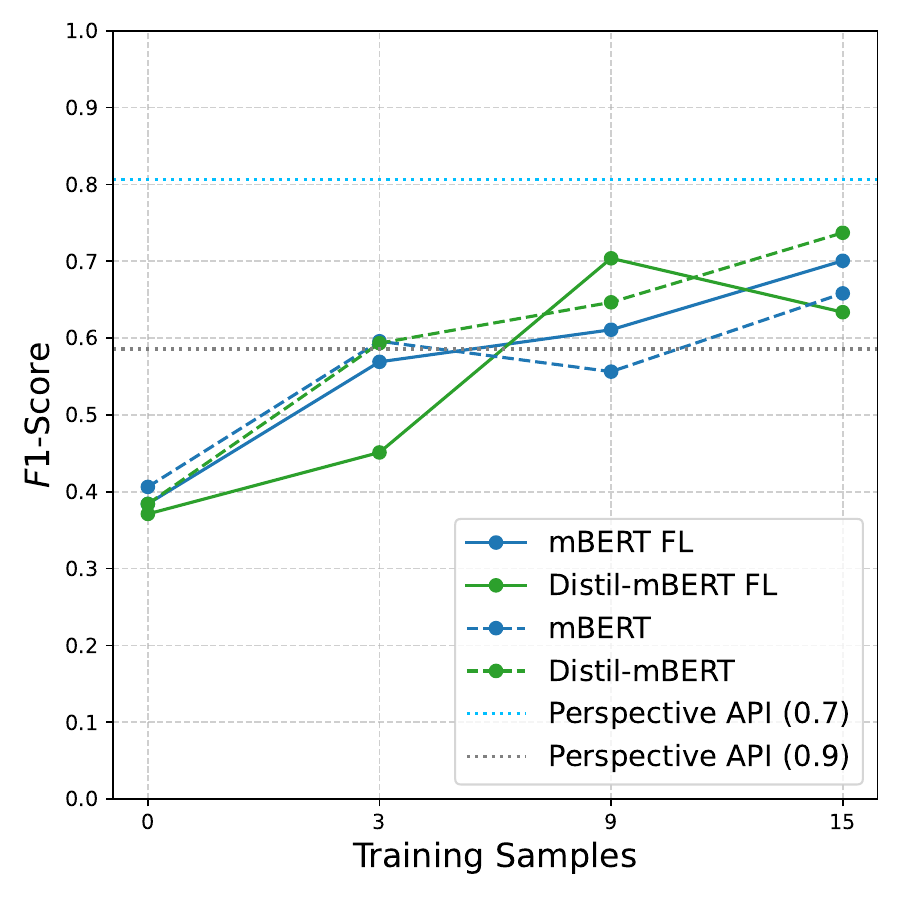}
        \caption{\texttt{rus-war}}
    \end{subfigure}
    \begin{subfigure}[b]{0.32\textwidth}
        \centering
        \includegraphics[width=\textwidth]{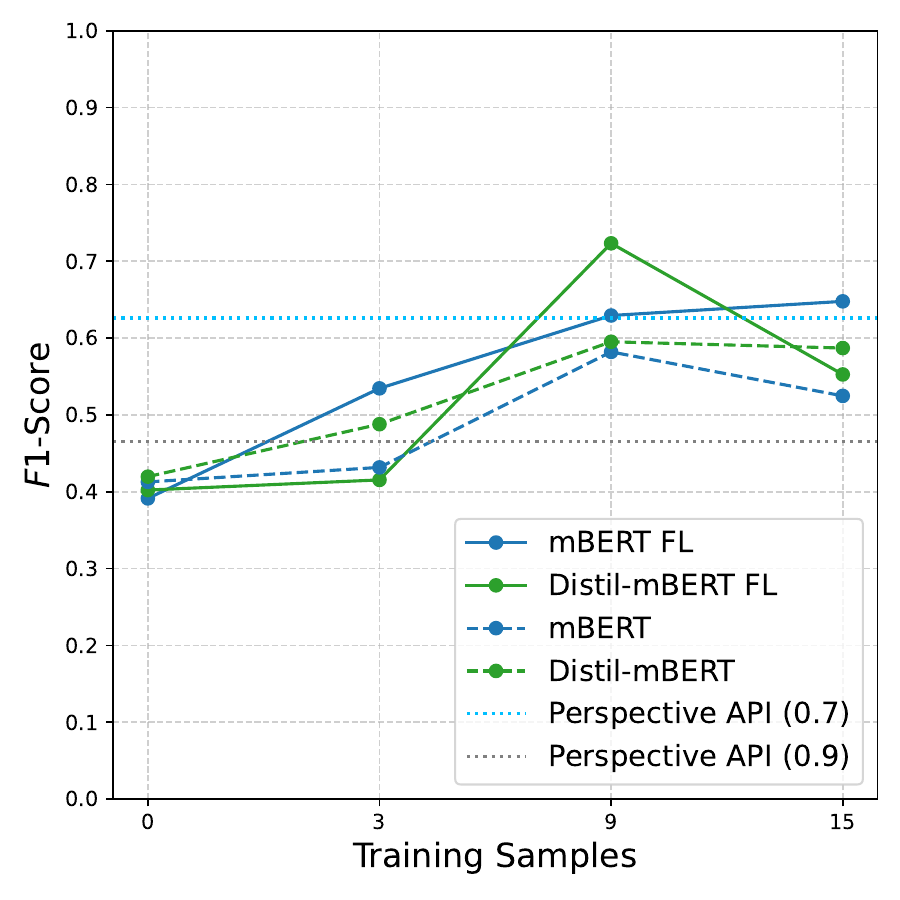}
        \caption{\texttt{server}}
    \end{subfigure}

    \caption{Comparison of $F_1$ scores using mBERT and Distil-mBERT across three training settings: FL (solid lines), single-target training (dashed lines), and Perspective API (horizontal dotted lines).
    Each subplot illustrates performance on a specific target group or the server.
    FL consistently improves client and server performance, especially with more (9-15) training samples.
    }
    \label{fig:baseline_vs_fl}
\end{figure*}

\begin{table*}
    \centering
    \resizebox{\textwidth}{!}{%
    \begin{tabular}{cc|rrrrrrrrrr}
    \toprule
        \multicolumn{1}{l}{} & Training & \multicolumn{2}{c}{\texttt{afr-black}} & \multicolumn{2}{c}{\texttt{afr-lgbtq}} & \multicolumn{2}{c}{\texttt{rus-lgbtq}} & \multicolumn{2}{c}{\texttt{rus-war}} & \multicolumn{2}{c}{\texttt{server}} \\
        \multicolumn{1}{l}{} & Samples & \multicolumn{1}{c}{M} & \multicolumn{1}{c}{D} & \multicolumn{1}{c}{M} & \multicolumn{1}{c}{D} & \multicolumn{1}{c}{M} & \multicolumn{1}{c}{D} & \multicolumn{1}{c}{M} & \multicolumn{1}{c}{D} & \multicolumn{1}{c}{M} & \multicolumn{1}{c}{D} \\
    \midrule
        \multirow{4}{*}{$\Delta$ No FL} & 0 & \textbf{0.04} & -0.01 & 0.00 & \underline{\textbf{0.02}} & \textbf{0.02} & -0.08 & -0.02 & -0.01 & -0.02 & -0.02 \\
         & 3 & -0.05 & -0.09 & 0.00 & -0.11 & \underline{\textbf{0.11}} & 0.00 & -0.03 & -0.14 & \textbf{0.10} & -0.07 \\
         & 9 & \textbf{0.02} & \underline{\textbf{0.06}} & \underline{\textbf{0.03}} & \textbf{0.01} & \textbf{0.04} & \underline{\textbf{0.03}} & \underline{\textbf{0.05}} & \underline{\textbf{0.06}} & \textbf{0.05} & \underline{\textbf{0.13}} \\
         & 15 & \underline{\textbf{0.15}} & \textbf{0.05} & \textbf{0.02} & -0.11 & \textbf{0.02} & -0.04 & \textbf{0.04} & -0.10 & \underline{\textbf{0.12}} & -0.03 \\
     \midrule
         & 0 & -0.07 & -0.10 & -0.05 & 0.00 & -0.37 & -0.38 & -0.42 & -0.44 & -0.23 & -0.22 \\
         $\Delta$ Perspective & 3 & \textbf{0.11} & \textbf{0.03} & \textbf{0.08} & -0.08 & -0.23 & -0.35 & -0.24 & -0.36 & -0.09 & -0.21 \\
         API (0.7) & 9 & \textbf{0.08} & \underline{\textbf{0.22}} & \textbf{0.03} & \underline{\textbf{0.22}} & -0.17 & -0.02 & -0.20 & -0.10 & 0.00 & \underline{\textbf{0.10}} \\
         & 15 & \underline{\textbf{0.17}} & \textbf{0.06} & \underline{\textbf{0.09}} & \textbf{0.03} & -0.08 & -0.05 & -0.11 & -0.17 & \underline{\textbf{0.02}} & -0.07 \\
    \midrule
         & 0 & -0.03 & -0.05 & \textbf{0.05} & \textbf{0.10} & -0.18 & -0.19 & -0.20 & -0.21 & -0.07 & -0.06 \\
         $\Delta$ Perspective & 3 & \textbf{0.15} & \textbf{0.07} & \textbf{0.18} & \textbf{0.01} & -0.04 & -0.16 & -0.02 & -0.13 & \textbf{0.07} & -0.05 \\
         API (0.9) & 9 & \textbf{0.12} & \underline{\textbf{0.26}} & \textbf{0.13} & \underline{\textbf{0.31}} & \textbf{0.02} & \underline{\textbf{0.17}} & \textbf{0.03} & \underline{\textbf{0.12}} & \textbf{0.16} & \underline{\textbf{0.26}} \\
         & 15 & \underline{\textbf{0.22}} & \textbf{0.10} & \underline{\textbf{0.19}} & \textbf{0.12} & \underline{\textbf{0.11}} & \textbf{0.14} & \underline{\textbf{0.11}} & \textbf{0.05} & \underline{\textbf{0.18}} & \textbf{0.09} \\
    \bottomrule
    \end{tabular}
    }
    \caption{$F_1$ differences between the three baseline settings and FL. \textbf{Bold}: FL improves the client performance. \underline{Underlined}: highest improvement for each setting and target group.
    M: mBERT, D: Distil-mBERT. mBERT benefits from FL with more data (15), whereas Distil-mBERT benefits the most with less data (9).
    }
    \label{tab:f1_diffs}
\end{table*}

\paragraph{RQ1: Performance of Perspective API varies}
As shown in Figure \ref{fig:baseline_vs_fl}, Perspective API performs strongly on Russian data, achieving $F_1$s of 0.75 and 0.81 for \texttt{rus-lgbtq} and \texttt{rus-war}, respectively, at the 0.7 threshold.
At the 0.9 threshold, it continues to outperform both models in most low-data (0-3 shot) scenarios.
However, its performance on Afrikaans, which it does not support, is notably poor and often fall below both FL and single-target fine-tuning.
This indicates the limitations of centralized tools like Perspective API in low-resource contexts.

\paragraph{RQ2: Individual clients benefit consistently from FL.}
Figure \ref{fig:baseline_vs_fl} compares classification results using FL (solid lines), single-target fine-tuning (dashed lines), and Perspective API (horizontal dotted lines), using both mBERT and Distil-mBERT.
Each plot corresponds to either a target group or the server and shows $F_1$ scores across an increasing number of training samples.
Table \ref{tab:f1_diffs} shows the $F_1$ improvements using FL over the baselines.
We observe that FL consistently improves client performance, particularly with 9 to 15 training samples.
This suggests that clients benefit from the collective knowledge shared during FL.
Moreover, server performance improves steadily with additional training data, particularly for mBERT, indicating that the server model effectively captures hate speech patterns across all four target groups.

\paragraph{RQ3: Personalization works, but performance varies.}

\begin{figure}[htbp]
    \centering
    \begin{subfigure}[b]{0.23\textwidth}
        \centering
        \includegraphics[width=\textwidth]{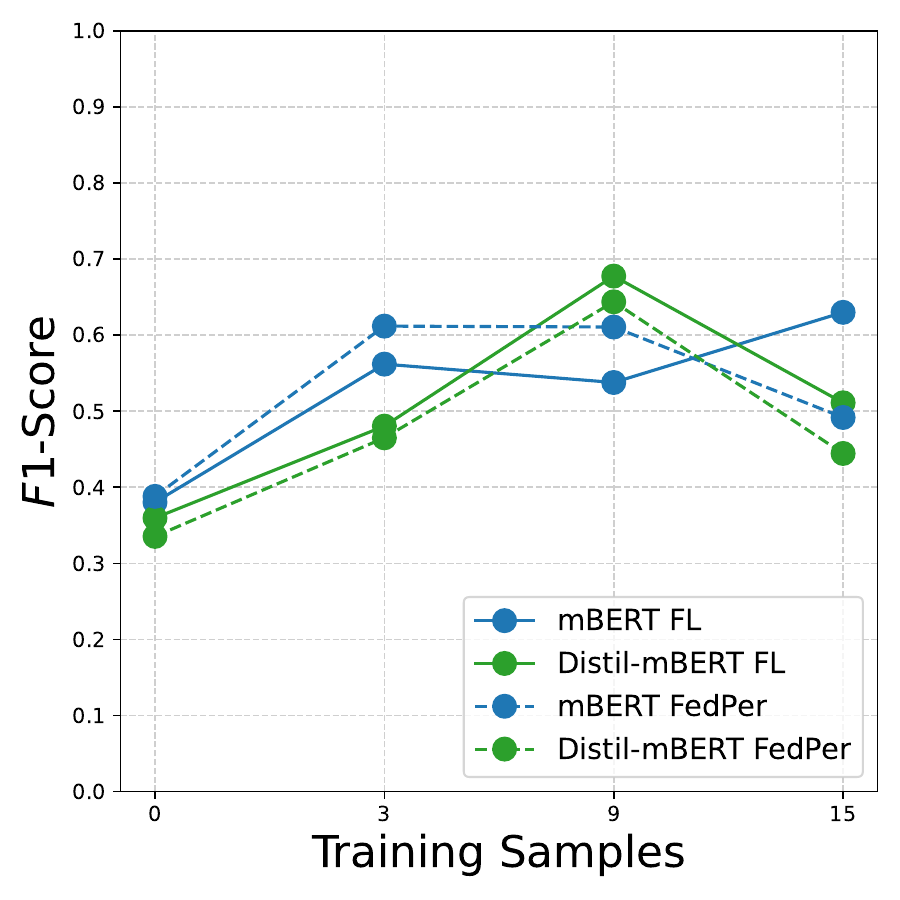}
        \caption{\texttt{afr-black}}
    \end{subfigure}
    \begin{subfigure}[b]{0.23\textwidth}
        \centering
        \includegraphics[width=\textwidth]{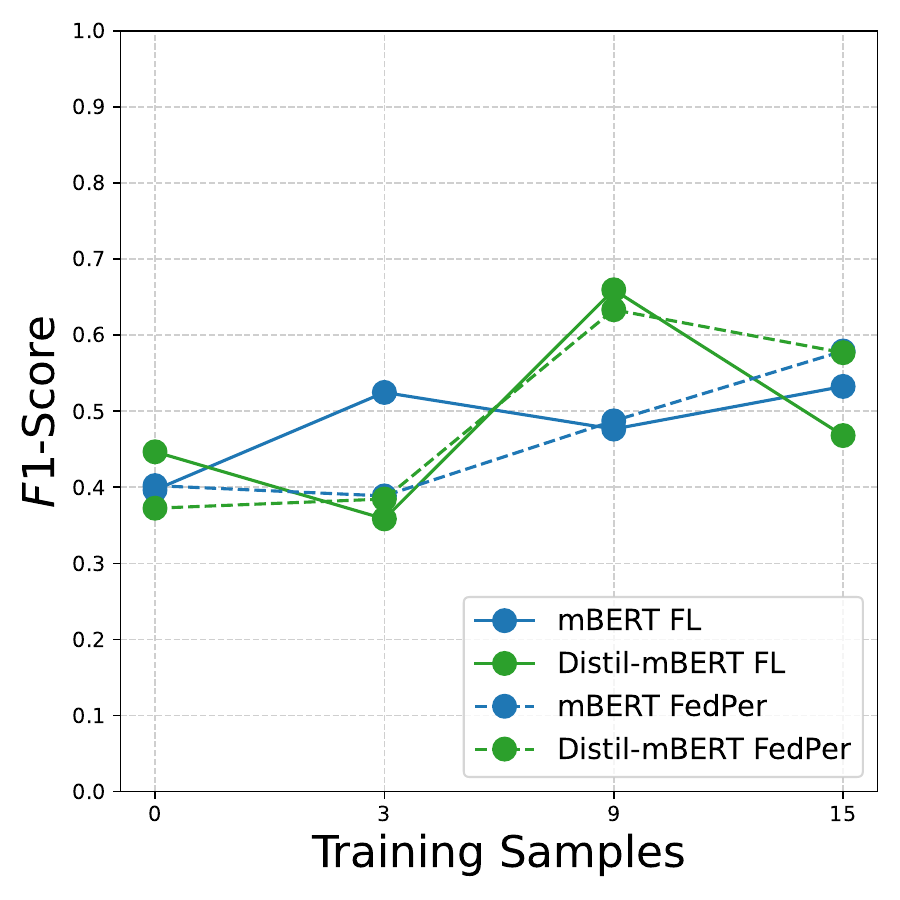}
        \caption{\texttt{afr-lgbtq}}
    \end{subfigure}

    \vspace{0.5cm}
    \begin{subfigure}[b]{0.23\textwidth}
        \centering
        \includegraphics[width=\textwidth]{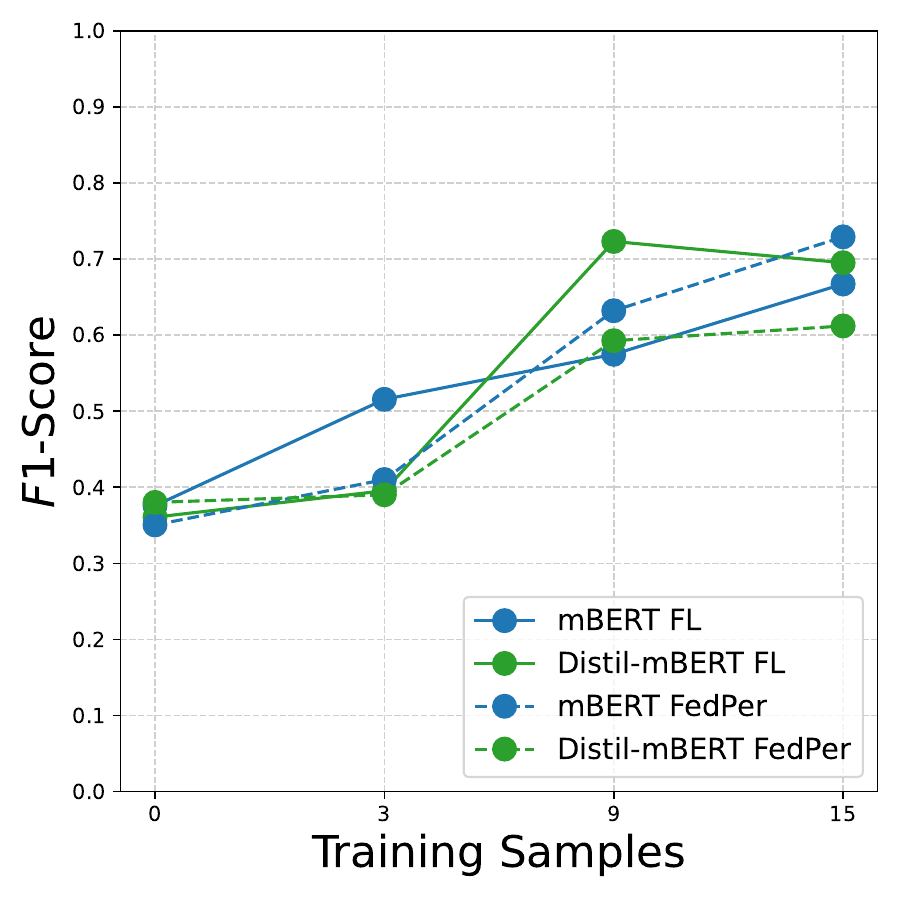}
        \caption{\texttt{rus-lgbtq}}
    \end{subfigure}
    \begin{subfigure}[b]{0.23\textwidth}
        \centering
        \includegraphics[width=\textwidth]{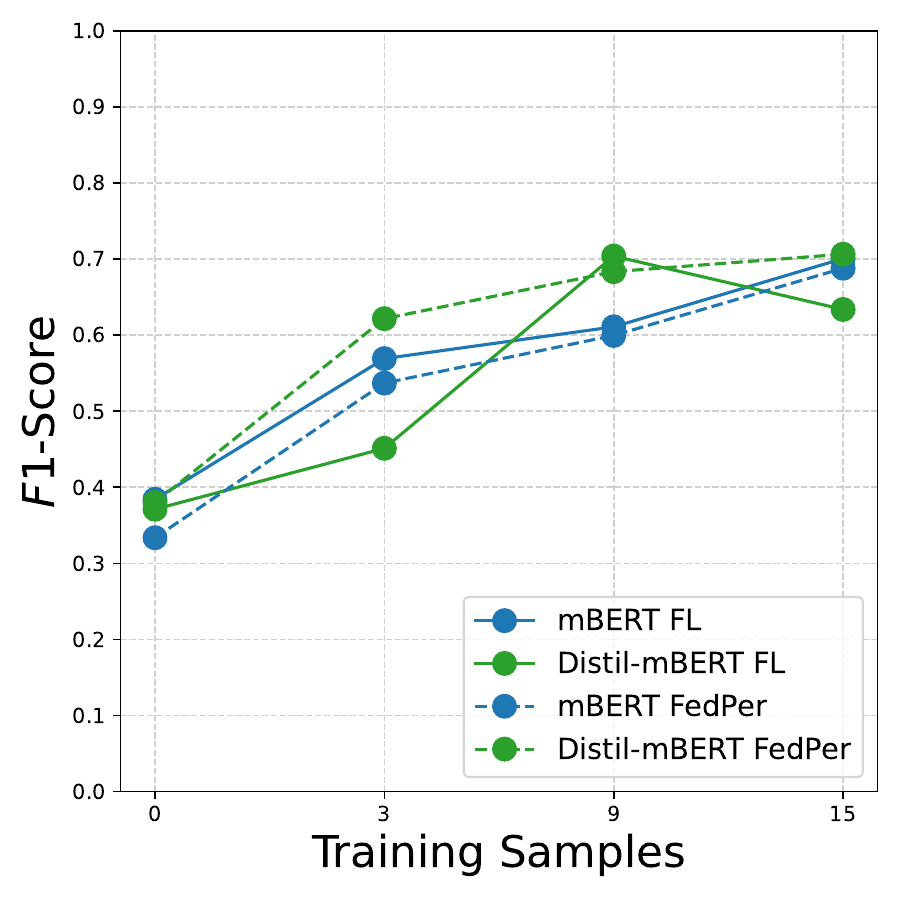}
        \centering
        \caption{\texttt{rus-war}}
    \end{subfigure}

    \caption{$F_1$ scores of client models customized using FedPer (dashed lines) are compared against those trained with standard FL (solid lines).
    Results are presented for the optimal $K_P$ value, which is 4 for both models.
    While FedPer occasionally yields modest improvements, its overall advantages are target- and language-specific.
    }
    \label{fig:fedper}
\end{figure}

\begin{figure}[htbp]
    \centering
    \begin{subfigure}[b]{0.23\textwidth}
        \centering
        \includegraphics[width=\textwidth]{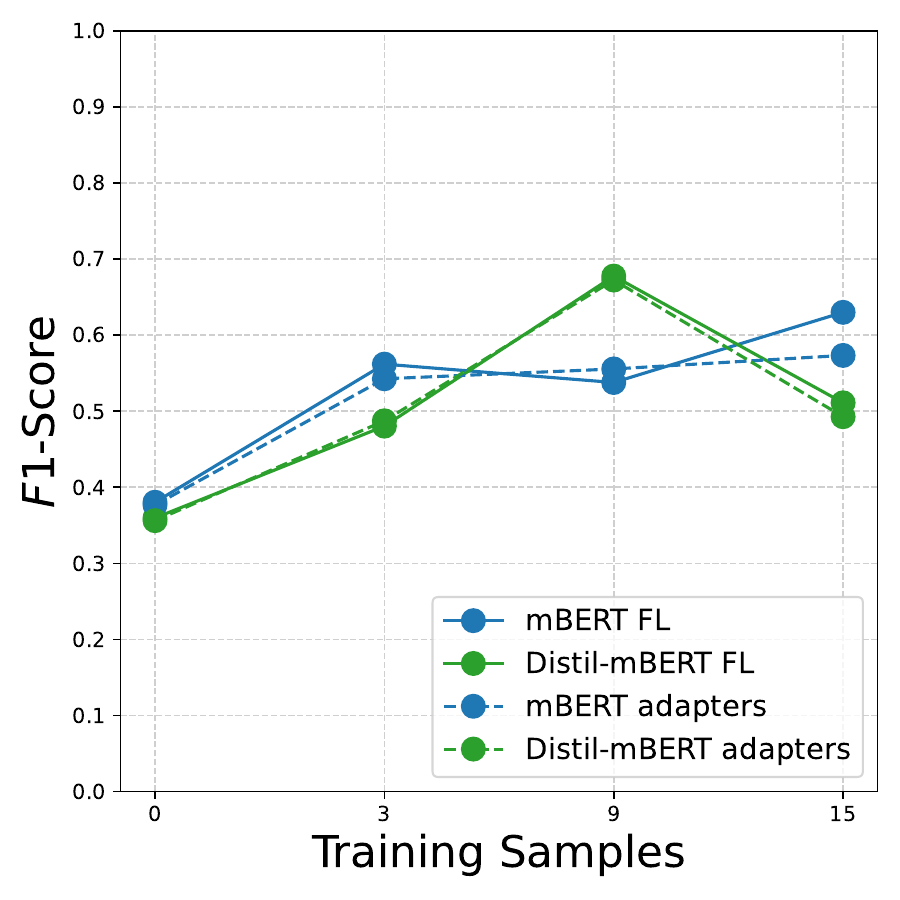}
        \caption{\texttt{afr-black}}
    \end{subfigure}
    \begin{subfigure}[b]{0.23\textwidth}
        \centering
        \includegraphics[width=\textwidth]{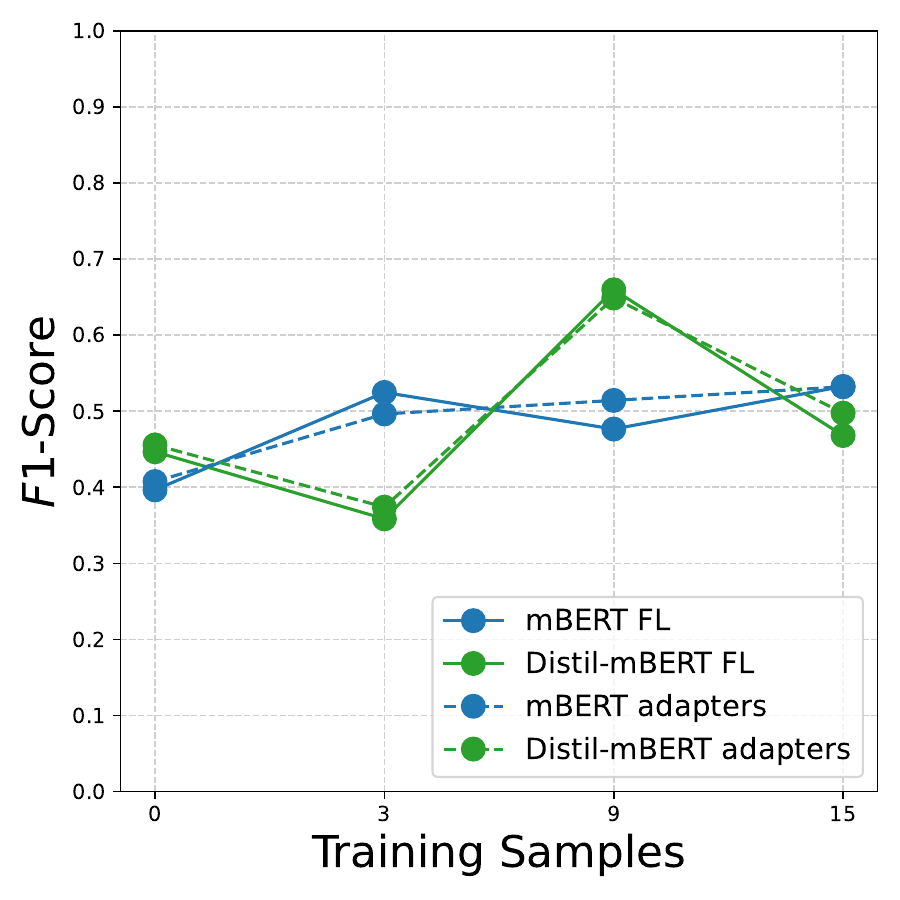}
        \caption{\texttt{afr-lgbtq}}
    \end{subfigure}

    \vspace{0.5cm}
    \begin{subfigure}[b]{0.23\textwidth}
        \centering
        \includegraphics[width=\textwidth]{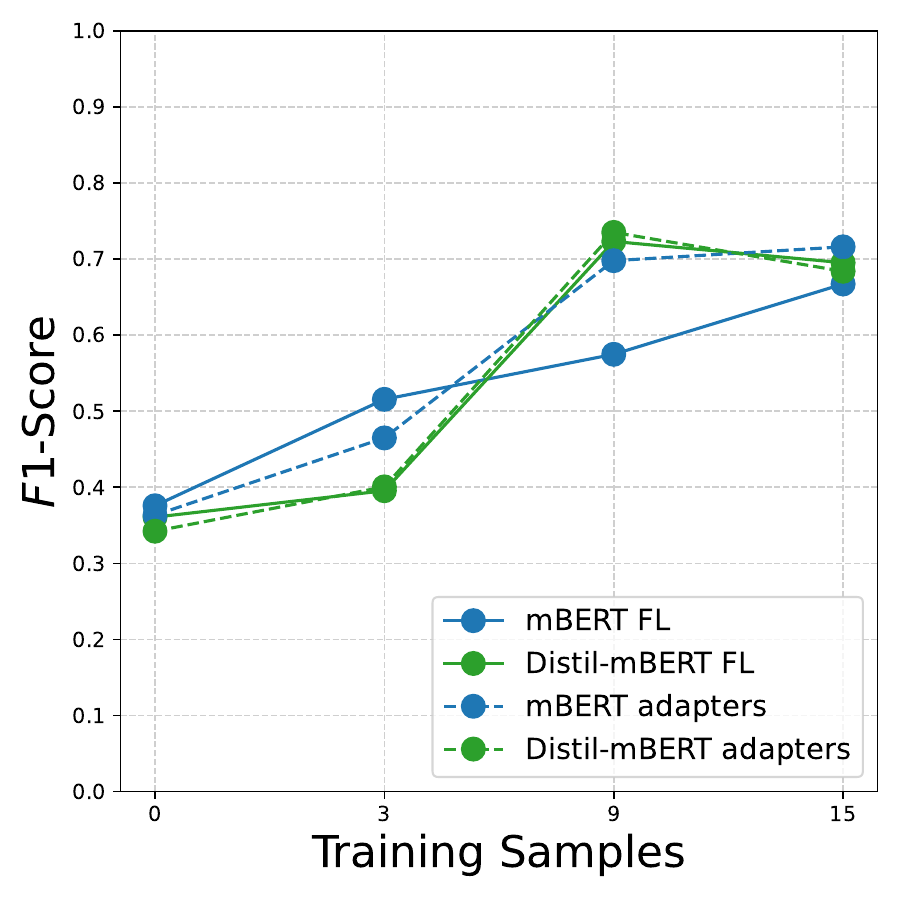}
        \caption{\texttt{rus-lgbtq}}
    \end{subfigure}
    \begin{subfigure}[b]{0.23\textwidth}
        \centering
        \includegraphics[width=\textwidth]{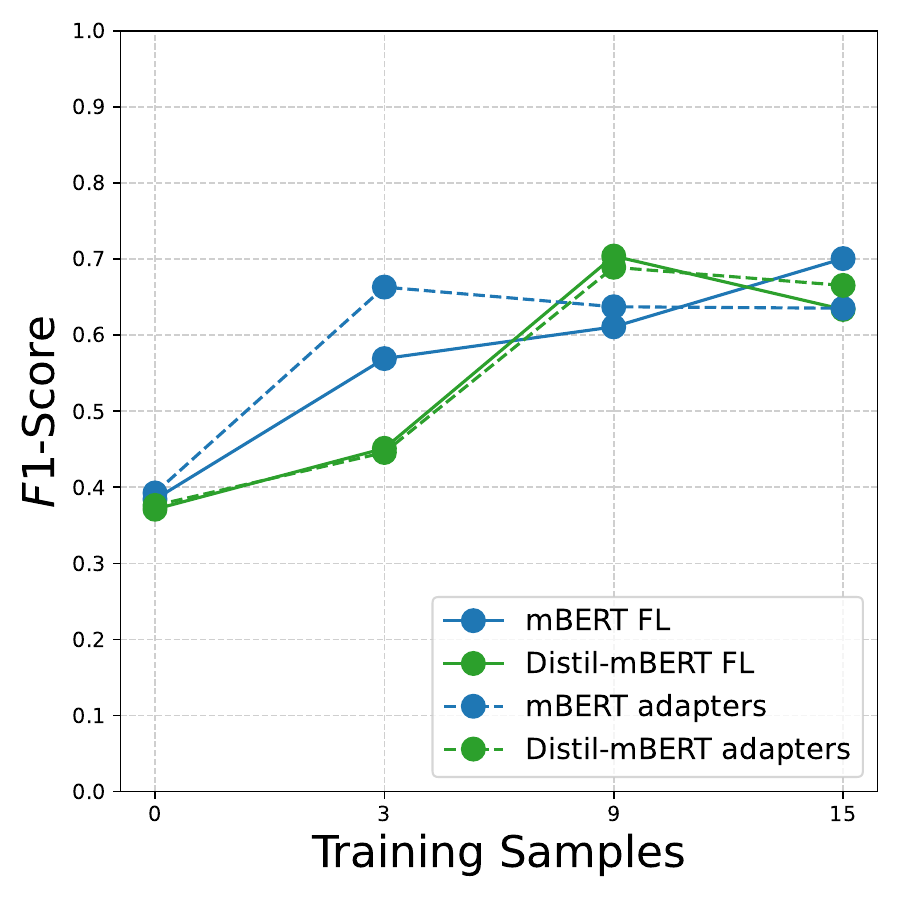}
        \centering
        \caption{\texttt{rus-war}}
    \end{subfigure}

    \caption{$F_1$ scores of client models customized using adapters
    and full-model fine-tuning (dashed lines), compared against those trained with standard FL (solid lines).
    Although a few clients see gains from adapter-based personalization, the overall improvement is unclear.
    }
    \label{fig:adapters}
\end{figure}

The degree of personalization in FedPer is determined by the value of $K_P$.
We test $K_P \in \{1, 2, 3, 4\}$ for both mBERT and Distil-mBERT, and report results using the best-performing $K_P$ for each model in Figure \ref{fig:fedper}.
Full results for all $K_P$ values are provided in \secref{fedper_full_results}.
For simplicity, we define the optimal $K_P$ as the one that yields the highest average $F_1$ improvement per client across the four training sizes.
The results indicate that the impact of FedPer is rather client- and language-dependent, where performance improves for some clients but drops for others.
For example, with mBERT and 15 training samples, \texttt{afr-black} suffers a sharp drop of 0.14 in $F_1$, whereas \texttt{rus-lgbtq} improves by 0.06.
Similar variability is observed with Distil-mBERT.
At 3-shot, all clients show performance declines (up to -0.16), yet all demonstrate improvements at 9-shot (up to 0.18).

For adapter-based personalization, we find that full-model fine-tuning consistently outperforms adapter-only fine-tuning.
Figure \ref{fig:adapters} presents full-model FL results with adapter personalization, and full results are shown in \secref{adapters_full_results}.
While certain clients, such as \texttt{rus-lgbtq} and \texttt{rus-war}, benefit from adapters (with mBERT gains of up to 0.13 and 0.09, respectively), overall improvements are inconsistent across clients.

\paragraph{Smaller models benefit slightly more from personalization.}
A comparison between standard FL (Figure \ref{fig:baseline_vs_fl}) and personalized FL results (Figures \ref{fig:fedper} and \ref{fig:adapters}) reveals that the smaller Distil-mBERT model benefits slightly more from FedPer than mBERT (an average $F_1$ improvement of 0.02 per client with the best-performing $K_P$).
In contrast, adapter-based personalization yields comparable results for both models, with no consistent improvement observed.

\section{Analysis}
\paragraph{Perspective API}
Since our data includes samples both with and without profanity, we expect the two chosen thresholds to influence the classification behavior of Perspective API.
We observe performance drops across all target groups when the threshold is raised from 0.7 to 0.9.
The difference is particularly pronounced in Russian, where the API otherwiese performs relatively well.
Increasing the threshold to 0.9 makes the API more conservative, reducing its sensitivity to hate.
While hateful sentences containing repeated profanity or highly offensive language are correctly identified under both thresholds, more subtle ones with little or no profanity are often missed at the higher threshold.
Simultaneously, the API is more reliant on profanity, more frequently correlating it with hate, as shown in \secref{api_analysis}.
Conversely, due to increased insensitivity to profanity, slightly profane yet positive sentences toward target groups, which are previously misclassified as hate, are correctly identified as non-hateful at the 0.9 threshold.

In addition to its threshold sensitivity, we find that Perspective API fails to detect culturally sensitive expressions, regardless of the threshold used.
For instance, ethnic slurs such as \foreignlanguage{russian}{хохлы} (\textit{Khokhols}) and \foreignlanguage{russian}{укры} (\textit{Ukry}), which are derogatory terms for Ukrainians, as well as homophobic slurs like \textit{Moffie} and \textit{skeef}, which are offensive references to effeminate or gay men, are not consistently flagged.
This is an indication that while Perspective API is effective for general-purpose hate speech detection, it lacks the cultural and linguistic nuance necessary for adaptation to specific cultural or ethnic contexts.

\paragraph{Effectiveness of personalization}
As shown by Figures \ref{fig:fedper} and \ref{fig:adapters}, both FedPer and adapters have variable effects on client models and are highly sensitive to the target group.
To assess their overall effectiveness, we compute the average $F_1$ improvement per client across all four training sizes.
While FedPer yields gains in specific cases, such as for \texttt{rus-war} using Distil-mBERT, Table \ref{tab:avg_improvement_personalization} shows that it does not consistently outperform non-personalized FL.
Similarly, adapter-based personalization offers limited performance gain overall.

Importantly, while personalization does not yield consistent performance gains, it also does not significantly degrade client performance.
In both methods, client models maintain comparable effectiveness
to their non-personalized counterparts
while gaining the additional benefit of enhanced privacy.
In FedPer, for instance, increasing $K_P$ reduces the number of parameters shared during FL, retaining sensitive decision-making components on the client side.

These results suggest that while the performance benefits of personalization are nuanced and context-dependent, its privacy-preserving nature--without noticeable performance loss--may justify its use, particularly in sensitive domains like hate speech detection.
Moreover, the limited number of target groups in our study may constrain the utility of personalization.
Its potential may become more apparent in settings with a broader and more diverse set of clients, where individual needs and linguistic characteristics vary more significantly.

\begin{table}
    \centering
    \begin{tabular}{l|rr}
        \toprule
        & \multicolumn{1}{c}{mBERT} & \multicolumn{1}{c}{Distil-mBERT} \\
        \midrule
        $K_P=1$ & -0.05 & -0.03 \\
        $K_P=2$ & -0.03 & -0.01 \\
        $K_P=3$ & -0.04 & -0.01 \\
        $K_P=4$ & -0.01 & 0.00 \\
        \midrule
        adapter-only & -0.13 & -0.10  \\
        full-model & 0.01 & 0.00 \\
        \bottomrule
    \end{tabular}
    \caption{Average $F_1$ improvement per client using FedPer with $K_P \in \{1, 2, 3, 4\}$ (top four rows) and two modes of adapter-based personalization (bottom two rows).
    }
    \label{tab:avg_improvement_personalization}
\end{table}


\section{Conclusion}
This work makes two key contributions.
First, we release \datasetname, a collection of localized and context-specific hate speech detection datasets.
\datasetname comprises data in four low-resource languages, covering six distinct target groups.
The datasets are curated by data collectors who are not only proficient in the target languages but also deeply familiar with the cultural nuances and contexts of hate speech in the respective countries.
Second, we evaluate the effectiveness of federated learning (FL), a privacy-preserving machine learning paradigm that keeps private data on user devices, in enabling few-shot hate speech detection using two lightweight multilingual models.
These models are suitable for deployment even on devices with limited computational resources.
We believe our findings will support future applications of privacy-aware hate speech filtering on user devices, such as a browser extension.

In addressing our research questions:
\textbf{(RQ1)} We find that both the Perspective API and zero-/few-shot learning with multilingual models perform reasonably well for detecting hate speech in the two tested low-resource languages.
\textbf{(RQ2)} Our results show modest but consistent improvements with FL under zero- and few-shot conditions (Figure \ref{fig:baseline_vs_fl}), highlighting its promise as a viable approach for privacy-preserving learning in low-resource settings, potentially applicable to other tasks.
\textbf{(RQ3)} Our investigation of two personalization methods reveals that their effectiveness is highly language- and target-dependent.
However, personalization offers a clear privacy advantage without significant performance loss.
We therefore see personalization as a promising direction, particularly in more resource-rich or heterogeneous environments.

\section*{Limitations}
Despite the comprehensive experimentation and valuable insights on federated hate speech detection presented in this study, several limitations remain, which we aim to address in future work.
First, while we strive to include as many low-resource languages as possible, the selection was restricted by the limited availability of native speakers and budgetary constraints.
This, in turn, limited the diversity and number of clients we could test.
Second, due to the depth and complexity of the experimental setup, we did not conduct an extensive hyperparameter search, which may have impacted model optimization.
Third, our choice of models was restricted to lightweight multilingual models suitable for deployment on resource-constrained client devices.
Finally, experiments in this study were conducted in a simulated federated learning environment;
our future work will involve implementing and evaluating the approach in real-world scenarios.

\section*{Ethics Statement}
In this work, we develop and utilize several hate speech detection datasets, the nature of which necessitates careful measures to protect data collectors from potential harm.
We ensure that data collectors are fully aware of the context of the target groups involved and obtain their consent for handling such data.
To minimize exposure to potentially harmful content, we randomly sample a small portion of the collected data for cross-annotation.
Additionally, data collectors are instructed to collect data exclusively from open domains to avoid copyright infringement and to remove any personally identifiable information, thereby maintaining the anonymity of the datasets.

While federated learning presents a promising approach to preserving user data privacy, it does not guarantee complete anonymity in the face of adversarial threats.
In certain circumstances, a malicious actor could potentially carry out attacks to infer personal information from data transmitted by individual clients, thus compromising the security of federated learning.
Therefore, additional precautions are recommended when implementing FL for sensitive data, with potential solutions including the application of differential privacy and the personalization of client models.


\bibliography{anthology,custom}
\bibliographystyle{acl_natbib}

\appendix

\section{Annotation details} \seclabel{annotation_details}

\subsection{Data collectors}
We recruit international students at
universities in the country of our research team
who are familiar with hate speech in the target countries as data collectors.
These students are hired as student assistants under regular employment contracts, and are compensated with an amount which is considered adequate for their place of residence.



\subsection{Instructions}
Prior to receiving the data collection prompt (\secref{react_dataset}), data collectors familiarize themselves with the six-category scheme using ``minimal pair'' examples with the same sentence differing only in polarity or the presence of profanity.
These examples provide a concrete understanding of subtle differences, such as the distinction between profane and non-profane expressions.
While these initial instructions offer a reference point, data collectors are free to interpret and define profanity based on the norms and nuances of their specific cultural contexts.

\subsection{Cross-annotation}
\begin{table}
    \centering
    \setlength{\tabcolsep}{3mm}{}
    \begin{tabular}{ll|r}
        \toprule
        language & target & \#sentences \\
        
        \midrule
        \multirow{2}{*}{Afrikaans} & Black people & 92 \\
                                   & LGBTQ & 352 \\
        \cmidrule{3-3}
        \multirow{2}{*}{Ukrainian} & Russians & 987  \\
                                   & Russophones & 1200 \\
        \cmidrule{3-3}
        \multirow{2}{*}{Russian} & LGBTQ & 115  \\
                                 & War victims & 193 \\
        \cmidrule{3-3}
        Korean & Women & 120 \\
        \bottomrule
    \end{tabular}
    \caption{
    The number of sentences in each cross-annotated dataset.
    }
    \label{tab:cross-annotation}
\end{table}

To ensure data quality, a subset of the data from all six categories is annotated by another native speaker of the language.
Details of the cross-annotated datasets are presented in Table \ref{tab:cross-annotation}.

\subsection{Inter-annotator agreement}
\begin{table}
    \centering
    \resizebox{\columnwidth}{!}{%
    \begin{tabular}{ll|rr|rr}
 \toprule
        language & target & \multicolumn{2}{c}{3 classes} & \multicolumn{2}{c}{2 classes} \\
         &  & $\kappa$ & $\alpha$ & $\kappa$ & $\alpha$ \\

        \midrule
        \multirow{2}{*}{Afrikaans} & Black people & 0.48 & 0.65 & 0.82 & 0.82 \\
                                   & LGBTQ & 0.57 & 0.71 & 0.58 & 0.57 \\
        \cmidrule{3-6}
        \multirow{2}{*}{Ukrainian} & Russians & 0.66 & 0.73 & 0.85 & 0.85  \\
                                   & Russophones & 0.47 & 0.70 & 0.86 & 0.86 \\
        \cmidrule{3-6}
        \multirow{2}{*}{Russian} & LGBTQ & 0.87 & 0.92 & 0.93 & 0.93  \\
                                 & War victims & 0.67 & 0.77 & 0.74 & 0.74 \\
        \cmidrule{3-6}
        Korean & Women & 0.66 & 0.80 & 0.60 & 0.60 \\
        \bottomrule
    \end{tabular}
    }
    \caption{Cohen's kappa ($\kappa$) and Krippendorff's alpha ($\alpha$) for the cross-annotated datasets.
    Values are shown for three classes (positive, neutral, hateful)
    and two classes (non-hateful and hateful).
    }
    \label{tab:interannotator}
\end{table}

We measure inter-annotator agreement using Cohen's kappa ($\kappa$) and Krippendorff's alpha ($\alpha$).
Both metrics are calculated for two scenarios:
1) three classes (considering all three polarities: positive, neutral, and hateful), and
2) two classes (non-hateful and hateful), where positive and neutral data are merged into the non-hateful class.
Table \ref{tab:interannotator} shows agreement scores for both metrics on each cross-annotated dataset.
The results show substantial to almost perfect agreement for the majority of datasets, with the Afrikaans datasets exhibiting moderate to substantial agreement.

\subsection{Corpus statistics}
We report corpus statistics for each \datasetname dataset in Table \ref{tab:corpus_statistics}.
These include the total number of sentences and tokens, the vocabulary size (unique tokens count), average, maximum, and minimum sentence lengths in tokens, standard deviation of sentence lengths, average word length in characters, type-token ratio, and the hapax legomena ratio.

\begin{table*}
    \centering
    \resizebox{\textwidth}{!}{%
    \begin{tabular}{l|rrrrrrr}
    \toprule
         & \multicolumn{1}{c}{afr-black} & \multicolumn{1}{c}{afr-lgbtq} & \multicolumn{1}{c}{ukr-russians} & \multicolumn{1}{c}{ukr-russophones} & \multicolumn{1}{c}{rus-lgbtq} & \multicolumn{1}{c}{rus-war} & \multicolumn{1}{c}{kor-women} \\
    \midrule
        \# Sentences & 2028 & 1016 & 1800 & 1200 & 772 & 1960 & 1294 \\
        \# Tokens & 34300 & 27647 & 26868 & 15283 & 11483 & 32566 & 14658 \\
        Vocab Size & 3754 & 4048 & 5363 & 3410 & 3441 & 7233 & 7018 \\
        Avg Sent Len (tok) & 16.91 & 27.45 & 14.93 & 12.74 & 15.09 & 16.62 & 11.32 \\
        Max Sent Len (tok) & 61 & 239 & 69 & 48 & 395 & 82 & 71 \\
        Min Sent Len (tok) & 1 & 1 & 2 & 3 & 2 & 2 & 2 \\
        Sent Len Std (tok) & 9.30 & 24.99 & 6.24 & 4.22 & 16.54 & 9.94 & 6.71 \\
        Avg Word Len (char) & 4.54 & 4.54 & 6.23 & 6.54 & 5.85 & 5.42 & 3.01 \\
        TTR & 0.11 & 0.15 & 0.20 & 0.22 & 0.30 & 0.22 & 0.48 \\
        Hapax Ratio & 0.01 & 0.08 & 0.11 & 0.14 & 0.15 & 0.08 & 0.36 \\
    \bottomrule
    \end{tabular}
    }
    \caption{
    Corpus statistics of the \datasetname datasets.
    }
    \label{tab:corpus_statistics}
\end{table*}

\section{AI-generated data} \label{sec:generated_data}
\subsection{Proportion of AI-generated data}
AI tools such as ChatGPT are employed to supplement data collection in cases where it is challenging to obtain sufficiently diverse examples in any of the three polarity categories.
Most of the AI-generated data falls under the positive category, where natural occurrences are considerably rarer compared to the neutral and negative categories.
Table \ref{tab:chatgpt_data} shows the proportion of AI-generated data within each dataset.

\begin{table}
    \centering
    \setlength{\tabcolsep}{3mm}{}
    \begin{tabular}{ll|r}
        \toprule
        language & target & generated data \\
        
        \midrule
        \multirow{2}{*}{Afrikaans} & Black people & 16.2\% \\ 
                                   & LGBTQ & 1.0\% \\ 
        \cmidrule{3-3}
        \multirow{2}{*}{Ukrainian} & Russians & 25.0\%  \\
                                   & Russophones & 35.0\% \\
        \cmidrule{3-3}
        \multirow{2}{*}{Russian} & LGBTQ & 19.6\% \\ 
                                 & War victims & 8.5\% \\ 
        \cmidrule{3-3}
        Korean & Women & 3.1\% \\ 
        \bottomrule
    \end{tabular}
    \caption{
    The proportion of AI-generated sentences (in percentage) within each dataset.
    }
    \label{tab:chatgpt_data}
\end{table}

\subsection{Prompts} \label{sec:prompts}
Following are some of the prompts to ChatGPT used to generate data.
\begin{itemize}
    \item Give me [\texttt{number}] neutral/positive sentences about [\texttt{target group}].
    \item Give me [\texttt{number}] positive or neutral sentences about [\texttt{target group}] in [\texttt{language}].
    \item Write positive/neutral/negative statements about [\texttt{target group}].
    \item I'm doing research to protect minority groups/[\texttt{target group}] and need [\texttt{number}] examples to add to my dataset.
    \item I'm searching for comments in [\texttt{language}] with the keyword [\texttt{target group}]. There are 6 categories: [...], could you search and give me some [\texttt{language}] comments with source URL and one of the categories?

\end{itemize}

\section{Model details}
\subsection{Models used} \seclabel{models_used}
To optimize the communication overhead between FL clients and the server, as well as allow models to be deployed on end devices with limited capacities, we focus on small language models for our study.
The following models have been used in our study, with the model sizes and number of layers shown:
\begin{itemize}
    \item XLM-RoBERTa (279M, 12 layers)\footnote{\url{https://huggingface.co/FacebookAI/xlm-roberta-base}}
    \item Multilingual BERT (179M, 12 layers)\footnote{\url{https://huggingface.co/google-bert/bert-base-multilingual-cased}}
    \item Multilingual DistilBERT (135M, 6 layers)\footnote{\url{https://huggingface.co/distilbert/distilbert-base-multilingual-cased}}
    \item DistilBERT (67M, 6 layers)\footnote{\url{https://huggingface.co/distilbert/distilbert-base-uncased}}
    \item Multilingual MiniLM (33M, 12 layers)\footnote{\url{https://huggingface.co/microsoft/Multilingual-MiniLM-L12-H384}}
    \item TinyBERT (14.5M, 4 layers)\footnote{\url{https://huggingface.co/huawei-noah/TinyBERT_General_4L_312D}}
    \item ALBERT (11.8M, 12 layers)\footnote{\url{https://huggingface.co/albert/albert-base-v2}}
\end{itemize}

\subsection{Model selection} 
\begin{figure*}[htbp]
    \centering
    \begin{subfigure}[b]{0.32\textwidth}
        \centering
        \includegraphics[width=\textwidth]{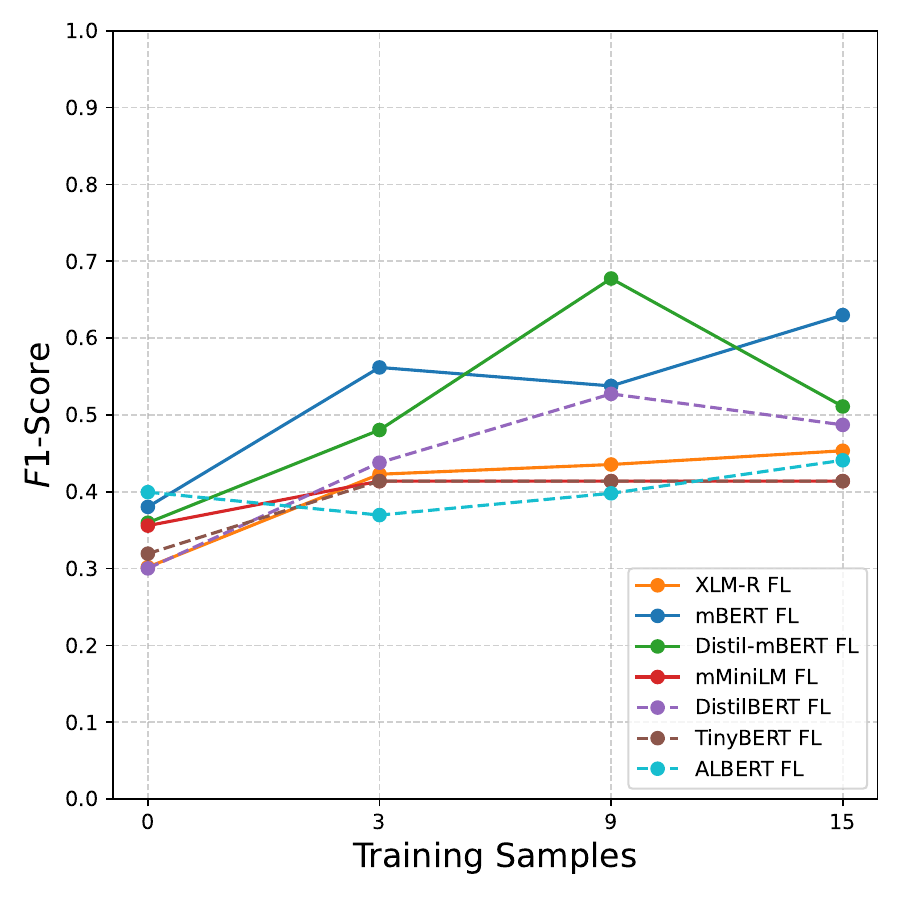}
        \caption{\texttt{afr-black}}
    \end{subfigure}
    \begin{subfigure}[b]{0.32\textwidth}
        \centering
        \includegraphics[width=\textwidth]{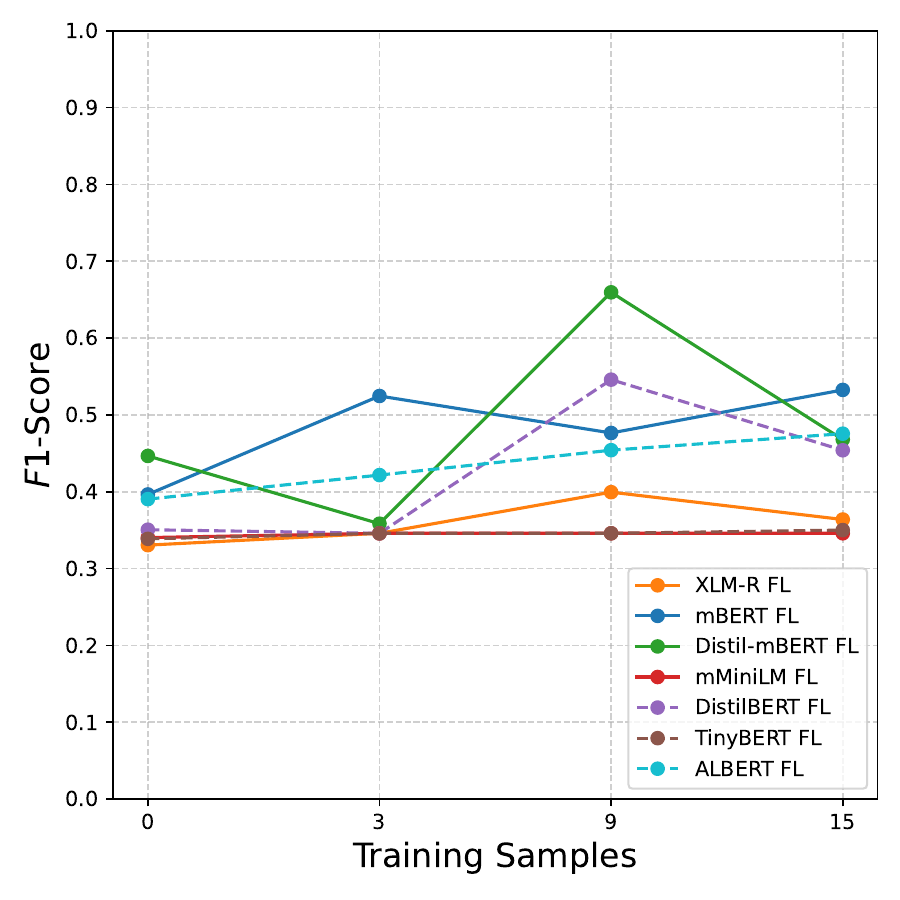}
        \caption{\texttt{afr-lgbtq}}
    \end{subfigure}
    \begin{subfigure}[b]{0.32\textwidth}
        \centering
        \includegraphics[width=\textwidth]{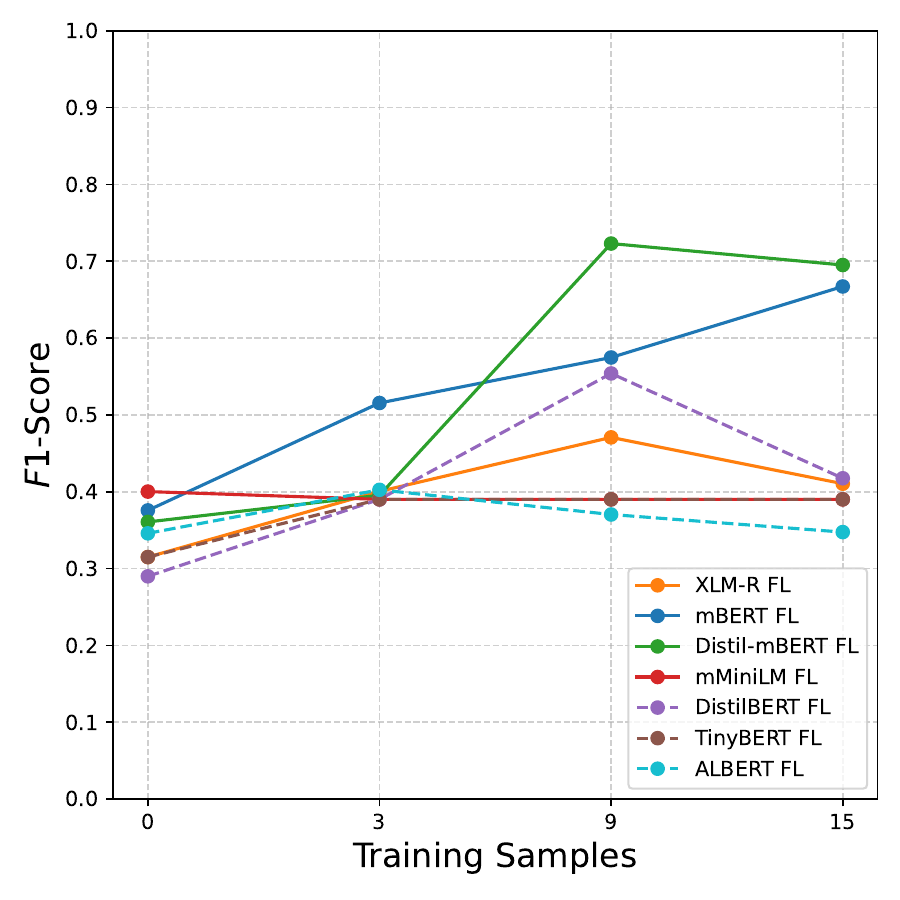}
        \caption{\texttt{rus-lgbtq}}
    \end{subfigure}

    \vspace{0.5cm}
    \begin{subfigure}[b]{0.32\textwidth}
        \centering
        \includegraphics[width=\textwidth]{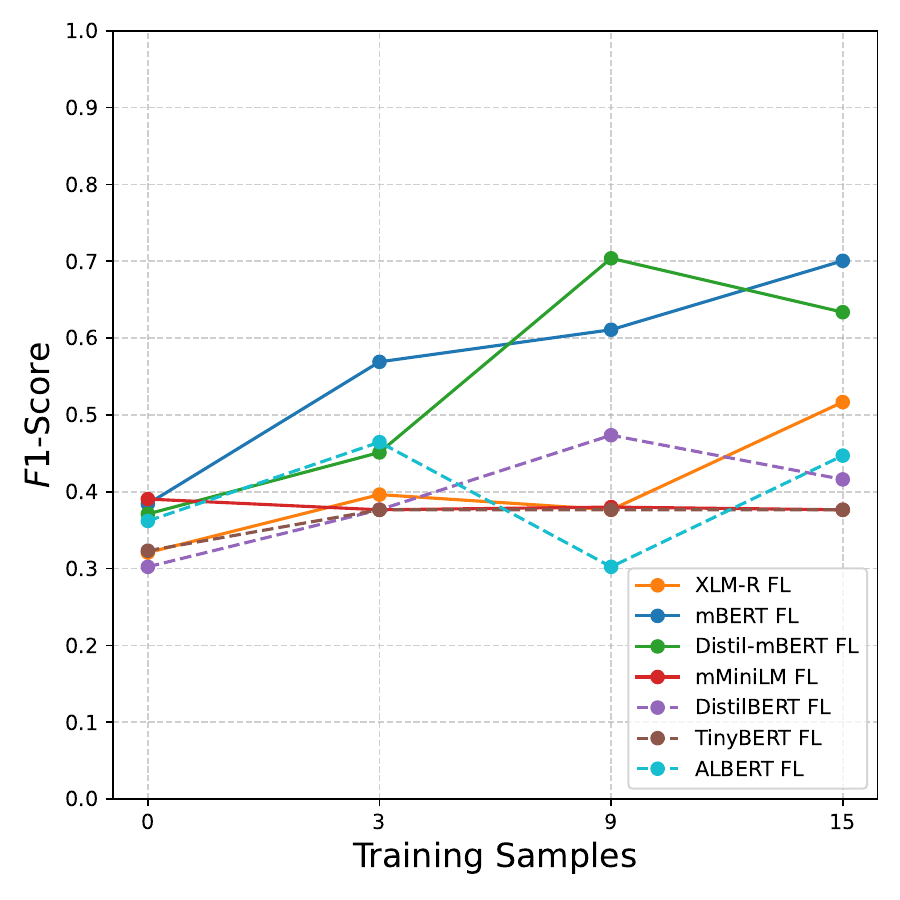}
        \caption{\texttt{rus-war}}
    \end{subfigure}
    \begin{subfigure}[b]{0.32\textwidth}
        \centering
        \includegraphics[width=\textwidth]{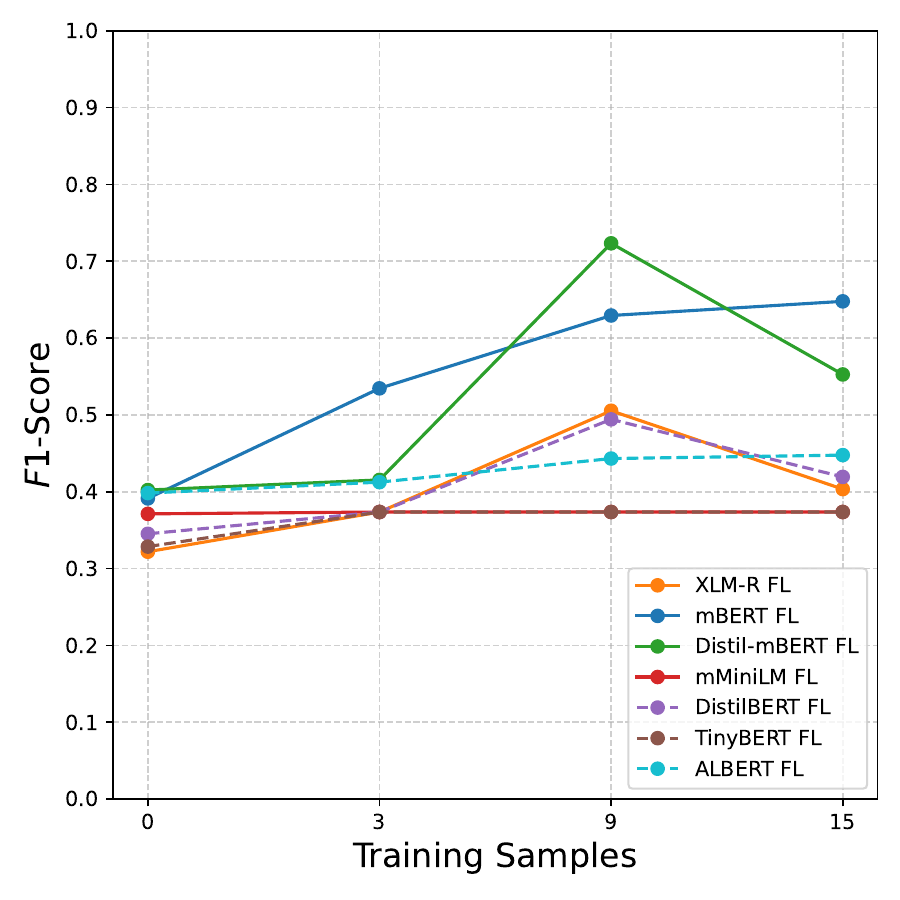}
        \caption{\texttt{server}}
    \end{subfigure}

    \caption{Comparison of $F_1$ scores of seven models, four multilingual and three monolingual.
    Each subplot shows performance on a specific target group or the server.
    The three monolingual models and multilingual MiniLM perform poorly across all target groups.
    Multilingual BERT and Distil-mBERT have the highest performance in most cases.
    }
    \label{fig:model_selection}
\end{figure*}

\seclabel{model_selection}
We evaluate the performance of the seven models in \secref{models_used} on classifying hate speech in a federated environment.
Four of the models are multilingual, the rest have not been explicitly trained on multilingual data.
Full results are shown in Figure \ref{fig:model_selection}.

\section{Selection of development and train data} \seclabel{dev_train_selection}
\begin{table}
    \centering
    \begin{tabular}{l|rrrr}
        & \multicolumn{1}{c}{\rotatebox{40}{afr-black}} 
        & \multicolumn{1}{c}{\rotatebox{40}{afr-lgbtq}} 
        & \multicolumn{1}{c}{\rotatebox{40}{rus-lgbtq}} 
        & \multicolumn{1}{c}{\rotatebox{40}{rus-war}} \\
        \toprule
        dev & 0.5 & 0.5 & 0.7 & 0.5 \\
        train&0.5 & 0.5 & 0.5 & 0.6 \\
    \end{tabular}
    \caption{Upper bounds of Levenshtein ratios for selecting development and train data.}
    \label{tab:levenshtein_ratio}
\end{table}

Because \datasetname exhibits potentially similar patterns due to its target-specificity, we mitigate possibly overlapping data by setting a threshold to the maximum Levenshtein ratio to accept a sentence when selecting development and train data.
By default, a Levenshtein ratio of $<$0.5 is used, meaning any sentence in the development set should have a Levenshtein similarity of less than 0.5 with any test data, and any sentence in the train set should have the same with any test or development data.
This ratio is slightly loosened in the case of \texttt{rus-lgbtq} and \texttt{rus-war} because the resulting datasets are too small.
In both cases, to ensure we do not include near-identical sentences accidentally, we sample sentences with a Levenshtein ratio of over 0.5 and manually check them against sentences they are reported to be similar with.
Table \ref{tab:dataset_sizes} presents the number of sentences in each split for the four target groups.

\begin{table}
    \centering
    \resizebox{\columnwidth}{!}{%
    \begin{tabular}{l|rrrr} \toprule
    
        & afr-black
        & afr-lgbtq
        & rus-lgbtq
        & rus-war \\
        \midrule
        train & 0-15 & 0-15 & 0-15 & 0-15 \\
        dev & 300 & 120 & 120 & 300 \\
        test & 87 & 225 & 111 & 154 \\ \bottomrule
    \end{tabular}
    }
    \caption{Number of sentences in the train, development, and test sets of each target group. We use 0, 3, 9, and 15 sentences per target group for training.}
    \label{tab:dataset_sizes}
\end{table}

\section{FedPer full results} 
\begin{figure*}[htbp]
    \centering
    \begin{subfigure}[b]{0.24\textwidth}
        \centering
        \includegraphics[width=\textwidth]{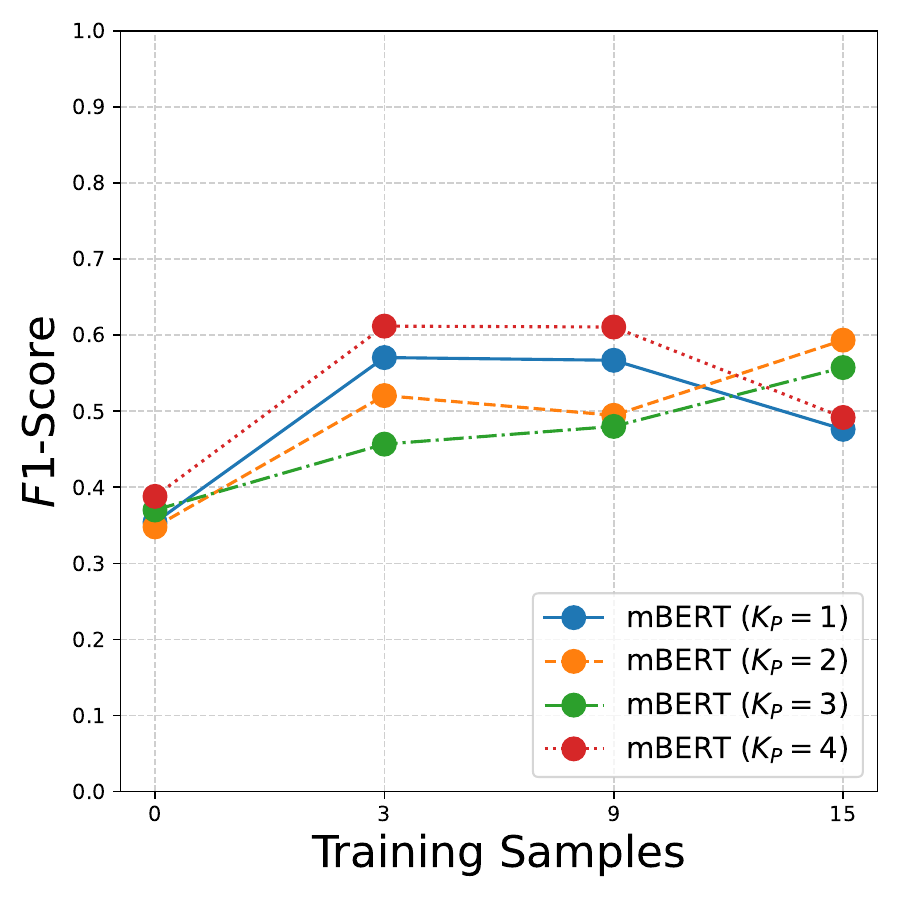}
        \caption{\texttt{afr-black}}
    \end{subfigure}
    \begin{subfigure}[b]{0.24\textwidth}
        \centering
        \includegraphics[width=\textwidth]{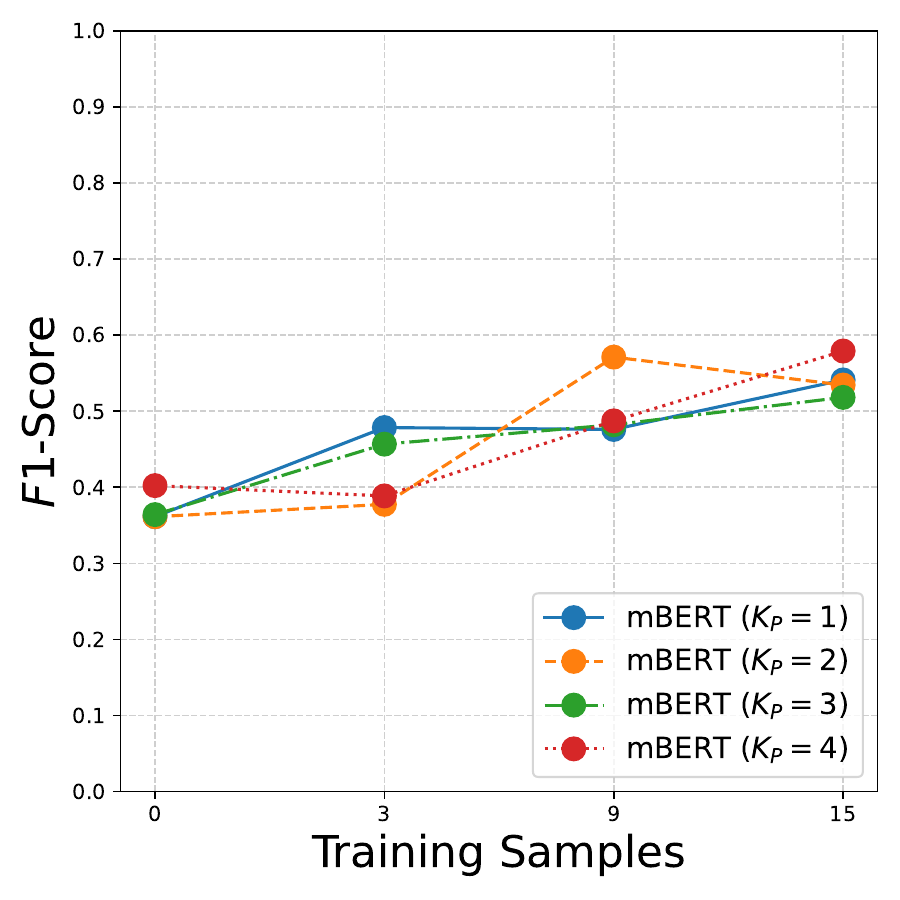}
        \caption{\texttt{afr-lgbtq}}
    \end{subfigure}
    \begin{subfigure}[b]{0.24\textwidth}
        \centering
        \includegraphics[width=\textwidth]{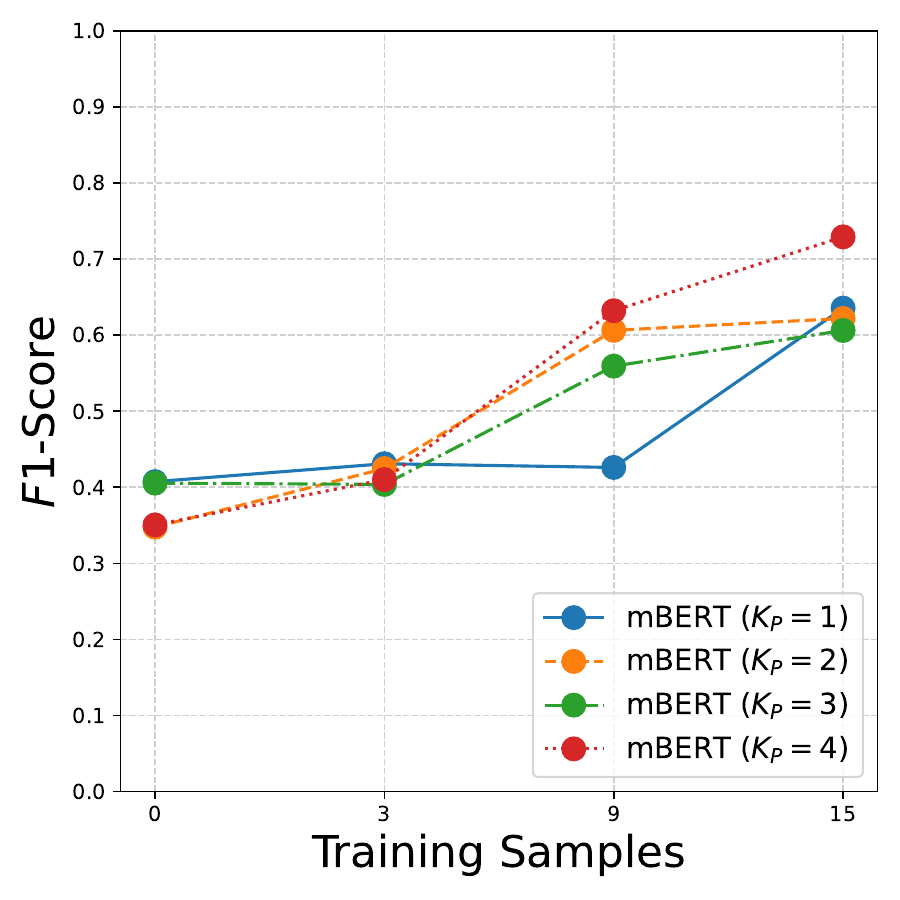}
        \caption{\texttt{rus-lgbtq}}
    \end{subfigure}
    \begin{subfigure}[b]{0.24\textwidth}
        \centering
        \includegraphics[width=\textwidth]{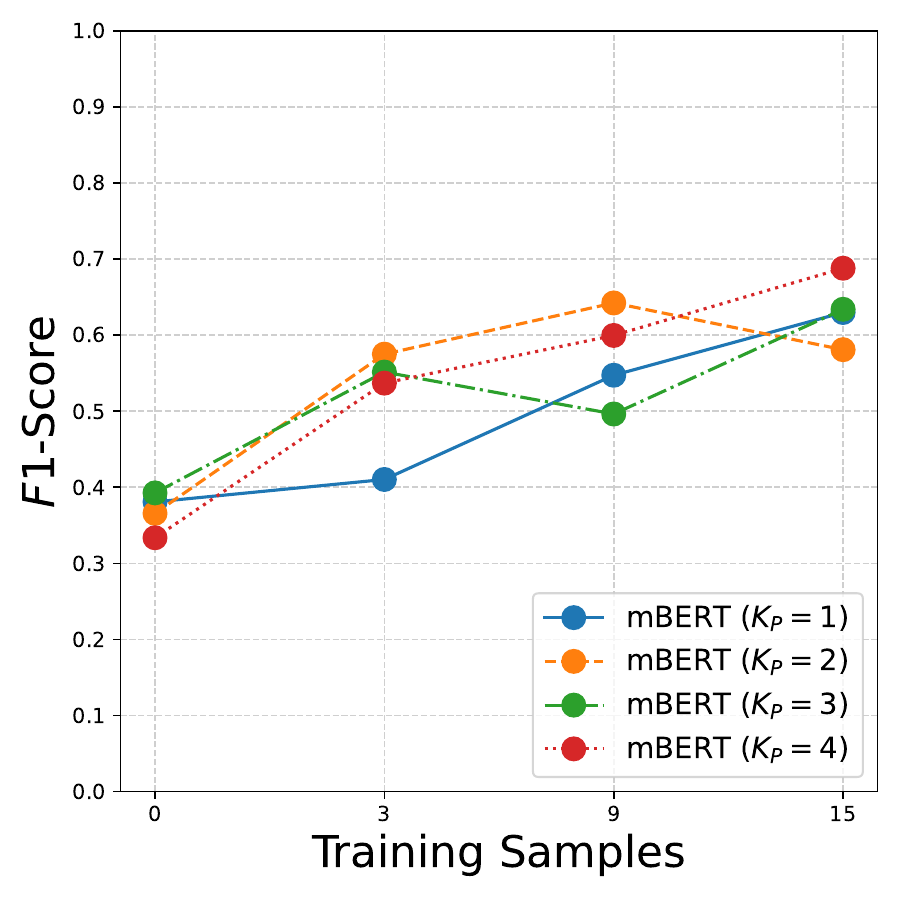}
        \caption{\texttt{rus-war}}
    \end{subfigure}

    \caption{FedPer results for mBERT. Each plot shows $F_1$ scores of a target group with $K_P$ (number of personalized layers) $\in \{1, 2, 3, 4\}$.}
    \label{fig:fedper_mbert_all}
\end{figure*}

\begin{figure*}[htbp]
    \centering
    \begin{subfigure}[b]{0.24\textwidth}
        \centering
        \includegraphics[width=\textwidth]{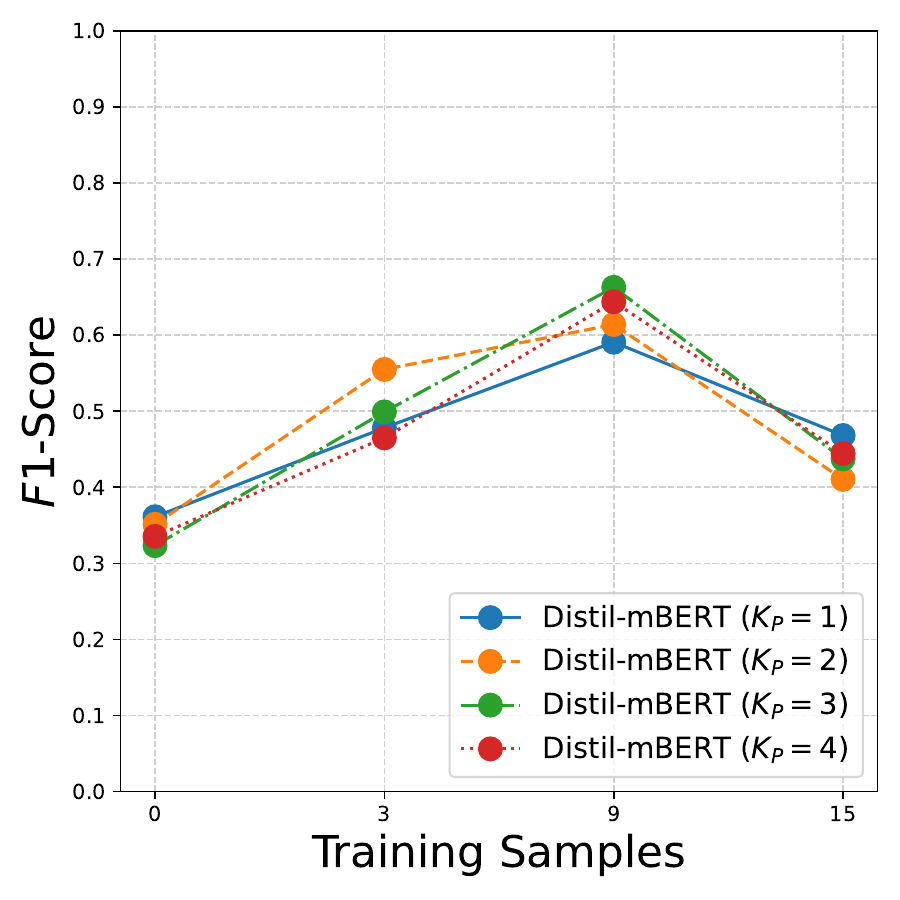}
        \caption{\texttt{afr-black}}
    \end{subfigure}
    \begin{subfigure}[b]{0.24\textwidth}
        \centering
        \includegraphics[width=\textwidth]{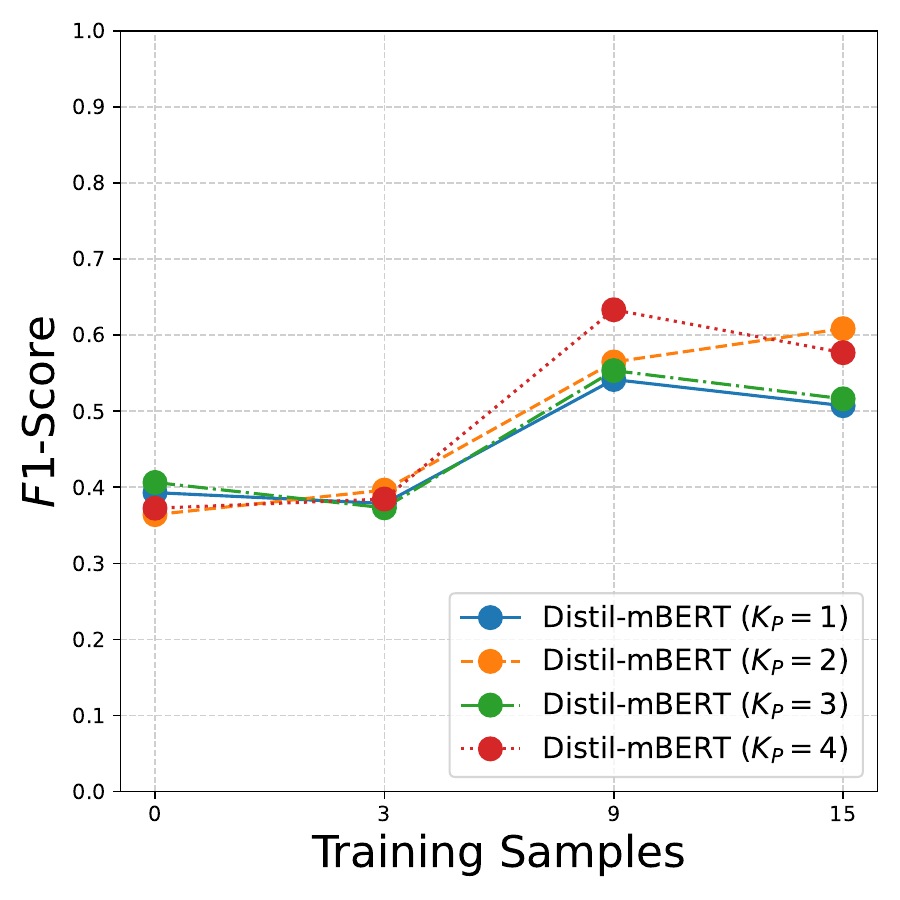}
        \caption{\texttt{afr-lgbtq}}
    \end{subfigure}
    \begin{subfigure}[b]{0.24\textwidth}
        \centering
        \includegraphics[width=\textwidth]{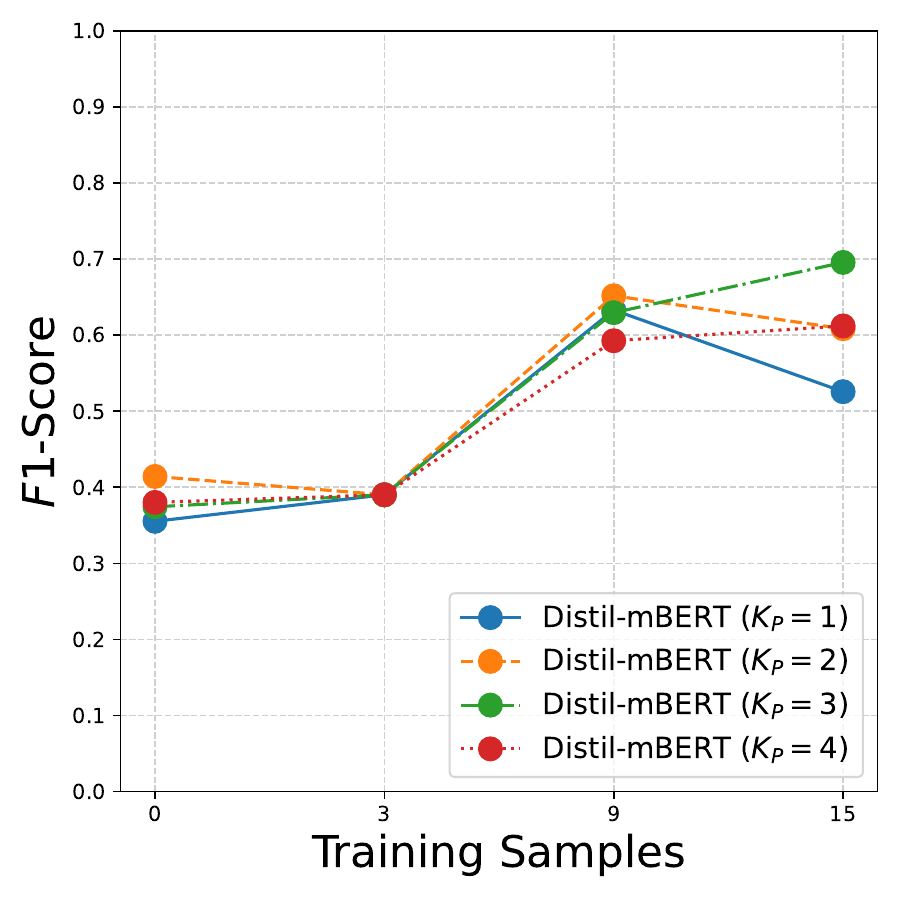}
        \caption{\texttt{rus-lgbtq}}
    \end{subfigure}
    \begin{subfigure}[b]{0.24\textwidth}
        \centering
        \includegraphics[width=\textwidth]{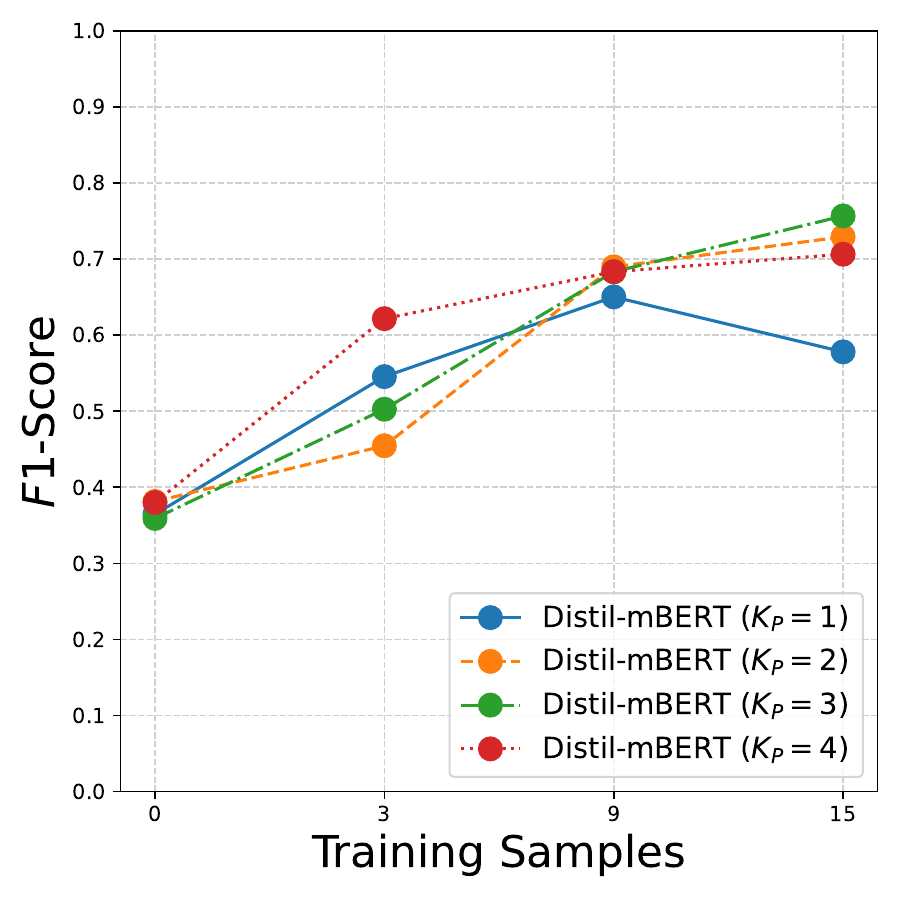}
        \caption{\texttt{rus-war}}
    \end{subfigure}

    \caption{FedPer results for Distil-mBERT. Each plot shows $F_1$ scores of a target group with $K_P$ (number of personalized layers) $\in \{1, 2, 3, 4\}$.}
    \label{fig:fedper_distil-mbert_all}
\end{figure*}

\seclabel{fedper_full_results}
We evaluate mBERT and Distil-mBERT using FedPer.
We test $K_P$ (number of personalized layers) values $\in \{1, 2, 3, 4\}$.
The complete results are shown in Figures \ref{fig:fedper_mbert_all}-\ref{fig:fedper_distil-mbert_all}.

\section{Adapters full results} 
\begin{figure*}[htbp]
    \centering
    \begin{subfigure}[b]{0.24\textwidth}
        \centering
        \includegraphics[width=\textwidth]{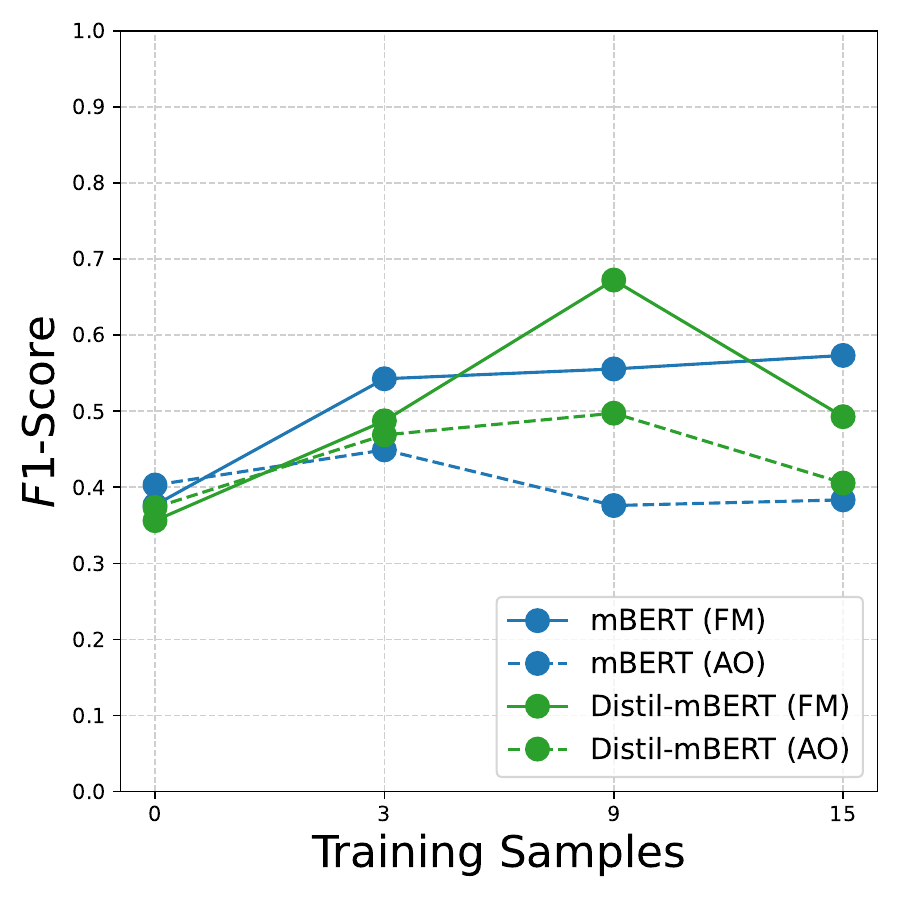}
        \caption{\texttt{afr-black}}
    \end{subfigure}
    \begin{subfigure}[b]{0.24\textwidth}
        \centering
        \includegraphics[width=\textwidth]{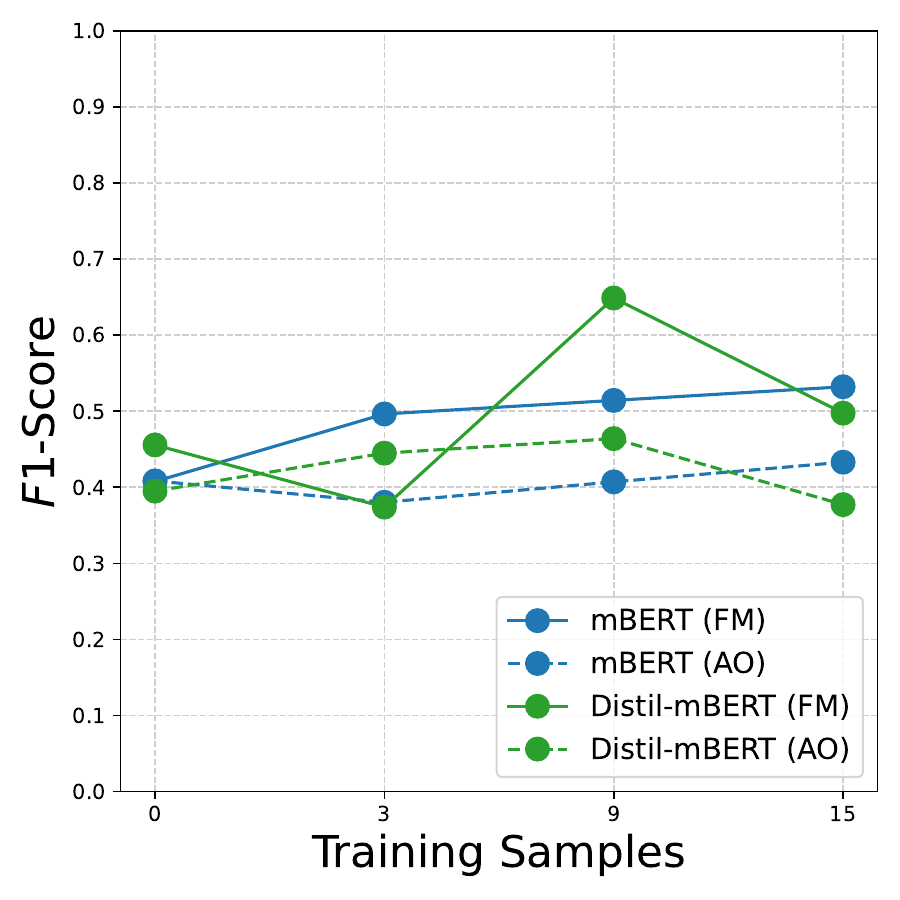}
        \caption{\texttt{afr-lgbtq}}
    \end{subfigure}
    \begin{subfigure}[b]{0.24\textwidth}
        \centering
        \includegraphics[width=\textwidth]{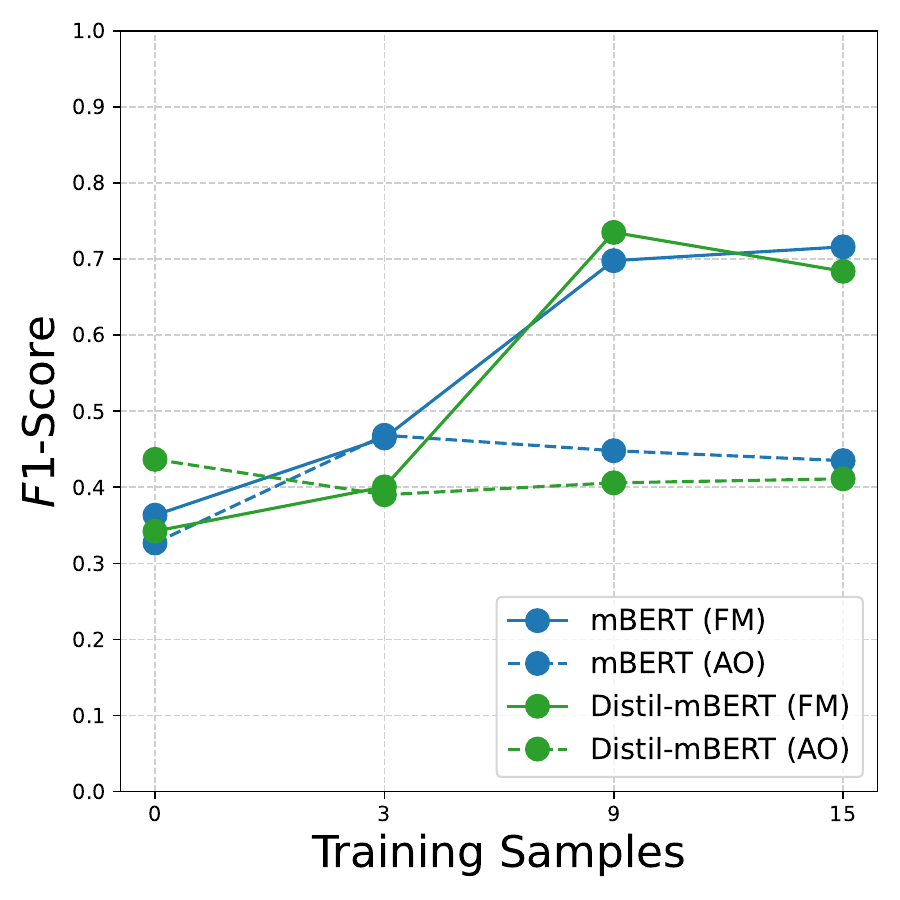}
        \caption{\texttt{rus-lgbtq}}
    \end{subfigure}
    \begin{subfigure}[b]{0.24\textwidth}
        \centering
        \includegraphics[width=\textwidth]{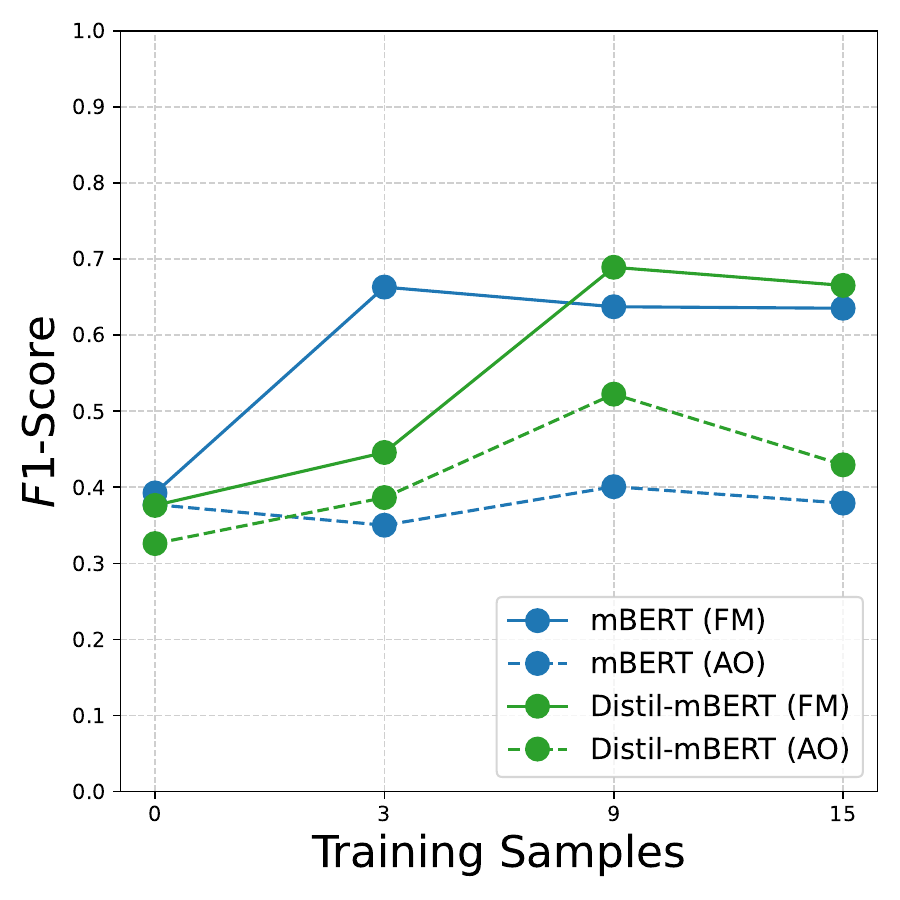}
        \caption{\texttt{rus-war}}
    \end{subfigure}

    \caption{
    Adapter-based personalization results for mBERT and Distil-mBERT.
    Results are compared between full-model fine-tuning (FM, solid lines) and adapter-only fine-tuning (AO, dashed lines).
    }
    \label{fig:adapter_full_results}
\end{figure*}

\seclabel{adapters_full_results}
We personalize client models by adding adapters and fine-tuning either the entire model, including the adapter parameters, or exclusively the adapter parameters. The complete evaluation results for mBERT and Distil-mBERT are shown in Figure \ref{fig:adapter_full_results}.

\section{Analysis of toxicity thresholds} \seclabel{api_analysis}
Table \ref{tab:thresholds} shows the percentages of sentences classified as hateful and non-hateful by Perspective API with thresholds 0.7 and 0.9, alongside the distribution in ground truth labels.
At both thresholds, Perspective API identifies substantially fewer hateful sentences (13.11\% and 3.44\%) compared to the ground truth (40.24\%), while simultaneously overestimating the proportion of non-hateful sentences.

While the ground truth data reflects a relatively balanced split between hateful sentences with (20.38\%) and without (19.86\%) profanity, Perspective API demonstrates a strong association between profanity and hate, shown by the higher proportions of profane sentences compared to non-profane ones among those classified as hateful.
This is especially pronounced at the 0.9 threshold, where 85.71\% of sentences labeled as hateful contain profanity, indicating a heavier reliance on profanity as a signal for hate compared to the 0.7 threshold.

\begin{table*}
    \centering
    \setlength{\tabcolsep}{4mm}{}
    \centering
    \begin{tabular}{lrrrrrr}
        \toprule
         & \multicolumn{2}{c}{API 0.7} & \multicolumn{2}{c}{API 0.9} & \multicolumn{2}{c}{Gold} \\
        \midrule
         & \multicolumn{1}{c}{P+} & \multicolumn{1}{c}{P-} & \multicolumn{1}{c}{P+} & \multicolumn{1}{c}{P-} & \multicolumn{1}{c}{P+} & \multicolumn{1}{c}{P-} \\
        \multicolumn{1}{r}{Hateful} & 9.34 & 3.77 & 2.95 & 0.49 & 20.38 & 19.86 \\
        \multicolumn{1}{r}{Not Hateful} & 40.98 & 45.90 & 47.38 & 49.18 & 27.53 & 32.23 \\
        \bottomrule
    \end{tabular}
    \caption{Percentages of sentences classified as ``Hateful'' and ``Not Hateful'' with (P+) and without (P-) profanity.
    API 0.7 (resp. 0.9): classified using Perspective API with threshold 0.7 (resp. 0.9). Gold: ground truth labels.}
    \label{tab:thresholds}
\end{table*}

\section{Examples of collected data} \seclabel{data_examples}
Table \ref{tab:data_examples} shows example sentences for each of the six categories in different languages.

As noted in \secref{react_dataset}, we occasionally adapt collected data to improve clarity with respect to the target group or intended polarity.
In some instances, we avoid the use of subjective slurs by replacing them with more neutral terms.
In the following positive example, the Russian term \foreignlanguage{russian}{хохлы} (\textit{Khokhols}), which is perceived as an ethnic slur by some, is replaced with the neutral term \foreignlanguage{russian}{украинцы} (\textit{Ukrainians}):
\begin{quote}
    \textbf{Original:} \foreignlanguage{russian}{Ну хохлы молодцы конечно блять.} (\textit{Well, the Khokhols sure did a good job, f*ck.})

    \textbf{Modified:} \foreignlanguage{russian}{Ну украинцы молодцы конечно блять.} (\textit{Well, the Ukrainians sure did a good job, f*ck.})
\end{quote}

In other cases, we remove subjective profanity to avoid introducing ambiguity in polarity, as demonstrated in the following neutral example:
\begin{quote}
    \textbf{Original:} \foreignlanguage{russian}{В Европе полно украинских беженцев, блять.} (\textit{There are tons of Ukrainian refugees in Europe, f*ck.})

    \textbf{Modified:} \foreignlanguage{russian}{В Европе полно украинских беженцев.} (\textit{There are tons of Ukrainian refugees in Europe.})
\end{quote}

We also occasionally add contextual information to clarify the intended polarity.
In the following sentence, additional information is provided to emphasize a positive stance:
\begin{quote}
    \textbf{Original:} \foreignlanguage{russian}{ЛГБТ+ добивается своего нахуй.} (\textit{LGBT+ are achieving what they f*cking want.})

    \textbf{Modified:} \foreignlanguage{russian}{ЛГБТ+ движение активно топит блять за права человека и добивается своего нахуй.} (\textit{The LGPT+ movement is actively pushing for human rights and is achieving what it f*cking want.})
\end{quote}

\begin{table*}
    \centering
    \begin{tabular}{cll}
        category & \multicolumn{1}{c}{text} & target group \\
        \toprule
        \multirow{2}{*}{P P+} & (afr) Swart Namibiërs is van die beste fokken mense ooit! & \multirow{2}{*}{Black people} \\
         & (eng) Black Namibians are some of the best fucking people out there! &  \\
        \midrule
        \multirow{2}{*}{P P-} & (ukr) \foreignlanguage{russian}{Росіяни хороші і не хочуть війни.} & \multirow{2}{*}{Russians} \\
         & (eng) Russians are good and do not want war. &  \\
        \midrule
        \multirow{2}{*}{N P+} & (kor) 여자들은 대부분 귀여운 것을 존나 좋아한다. & \multirow{2}{*}{Women} \\
         & (eng) Most women are fucking fond of cute things. &  \\
        \midrule
        \multirow{2}{*}{N P-} & (rus) \foreignlanguage{russian}{Беженцы из Украины рассказывают о жизни в оккупации.} & \multirow{2}{*}{War victims} \\
         & (eng) Refugees from Ukraine talk about life under occupation. &  \\
        \midrule
        \multirow{2}{*}{H P+} & (ukr) \foreignlanguage{russian}{Скільки ви ще будете хрюкати, уроди російськомовні?!} & \multirow{2}{*}{Russophones} \\
         & (eng) How much longer will you grunt, you Russian-speaking freaks?! &  \\
        \midrule
        \multirow{2}{*}{H P-} & (afr) Daar is nie plek vir homoseksuele in Namibië nie. & \multirow{2}{*}{LGBTQ} \\
         & (eng) There is no place for homosexuals in Namibia. &  \\
    \end{tabular}
    \caption{Example data for each category. The first part of the category name indicates the polarity (P: positive, N: neutral, H: hateful). The second part indicates the presence of profanity (P+: with profanity, P-: without profanity).}
    \label{tab:data_examples}
\end{table*}

\section{Computation}
Standard FL and FedPer experiments using mBERT and Distil-mBERT with four FL clients can be completed in 20-30 minutes using four NVIDIA GeForce RTX 2080 Ti GPUs.
Adapter-personalized FL experiments with the same client models can be completed in about 30 minutes on four NVIDIA RTX A6000 GPUs.

\end{document}